%% file: camera-ready.tex
\documentclass{article} % For LaTeX2e
\usepackage{iclr2021_conference,times}

\input{math_commands.tex}

\usepackage{url}

\usepackage{booktabs}       % professional-quality tables
\usepackage{amsfonts}       % blackboard math symbols
\usepackage{nicefrac}       % compact symbols for 1/2, etc.
\usepackage{microtype}      % microtypography
\usepackage{makecell}
\usepackage{multirow}
\usepackage{amsmath}
\usepackage{amssymb}
\usepackage{graphicx}
\usepackage{wrapfig}
\usepackage{lipsum}
\usepackage{subcaption}
\definecolor{citecolor}{RGB}{34, 149, 34}
\usepackage[pagebackref=true,breaklinks=true,letterpaper=true,colorlinks,citecolor=citecolor,bookmarks=false]{hyperref}

\usepackage{epsfig}
\usepackage{kotex}
\usepackage{booktabs}
\usepackage{multirow}
\usepackage{dsfont}
\usepackage{xcolor}
\usepackage{subcaption}
\usepackage{pifont}
\usepackage{algorithm,algorithmicx,algpseudocode}

\usepackage{amsmath}
\usepackage{amssymb}
\usepackage{bm}
\usepackage{wrapfig}

\usepackage{pifont}
\usepackage{colortbl}

\usepackage{booktabs}
\usepackage{tabulary}
\usepackage{soul}
\usepackage{xspace}\xspace
\usepackage{adjustbox}
\usepackage{array}

\usepackage{dblfloatfix}
\usepackage{transparent}

\usepackage{listings}
\usepackage{caption}

\definecolor{darkergreen}{RGB}{21, 152, 56}
\definecolor{red2}{RGB}{252, 54, 65}

\newcolumntype{x}[1]{>{\centering\arraybackslash}p{#1pt}}
\newcolumntype{y}[1]{>{\raggedright\arraybackslash}p{#1pt}}
\newcolumntype{z}[1]{>{\raggedleft\arraybackslash}p{#1pt}}
\newlength\savewidth

\makeatletter\renewcommand\paragraph{\@startsection{paragraph}{4}{\z@}
  {.5em \@plus1ex \@minus.2ex}{-.5em}{\normalfont\normalsize\bfseries}}\makeatother

\definecolor{mygray}{gray}{.9}

\title{Is Label Smoothing Truly Incompatible with Knowledge Distillation: An Empirical Study}

\iclrfinalcopy

\author{Zhiqiang Shen  \\
CMU\\
\And
Zechun Liu  \\
CMU \& HKUST \\
\And
Dejia Xu  \\
Peking University \\
\And
Zitian Chen  \\
UMass Amherst \\
\And
Kwang-Ting Cheng  \\
HKUST \\
\And
Marios Savvides \\
CMU \\
}

\begin{document}

\maketitle

\begin{abstract}
This work aims to empirically clarify a recently discovered perspective that label smoothing is incompatible with knowledge distillation~\citep{muller2019does}. We begin by introducing the  motivation behind on how this incompatibility is raised, i.e., {\em label smoothing erases relative information between teacher logits}. We provide a novel connection on how label smoothing affects distributions of semantically similar and dissimilar classes. Then we propose a metric to quantitatively measure the degree of erased information in sample's representation. After that, we study its one-sidedness and imperfection of the incompatibility view through massive analyses, visualizations and comprehensive experiments on Image Classification, Binary Networks, and Neural Machine Translation. 
Finally, we broadly discuss several circumstances wherein label smoothing will indeed lose its effectiveness.\footnote{Project page: \url{http://zhiqiangshen.com/projects/LS_and_KD/index.html}.}
\end{abstract}

\section{Introduction}

Label smoothing~\citep{szegedy2016rethinking} and knowledge distillation~\citep{hinton2015distilling} are two commonly recognized techniques in training deep neural networks and have been applied in many state-of-the-art models, such as language translation~\citep{vaswani2017attention,tan2018multilingual,Zhou2020Understanding}, image classification~\citep{xie2019self,he2019bag} and speech recognition~\citep{chiu2018state,pereyra2017regularizing,chorowski2017towards}. Recently a large body of studies is focusing on exploring the underlying relationships between these two methods, for instance, M{\"u}ller et al.~\citep{muller2019does} discovered that label smoothing could improve calibration implicitly but will hurt the effectiveness of knowledge distillation. Yuan et al.~\citep{yuan2019revisit} considered knowledge distillation as a dynamical form of label smoothing as it delivered a regularization effect in training. The recent study~\citep{lukasik2020does} further noticed label smoothing could help mitigate label noise, they showed that when distilling models from noisy data, the teacher with label smoothing is helpful. Despite the massive and intensive researches, how to use label smoothing as well as knowledge distillation in practice is still unclear, divergent, and under-explored. Moreover, it is hard to answer when and why label smoothing works well or not under a variety of discrepant circumstances.

\vspace{-0.05in}
\noindent{\textbf{View of incompatibility between label smoothing and knowledge distillation.}}
Recently, M{\"u}ller et al. proposed a new standpoint that {\em teachers trained with label smoothing distill inferior student compared to teachers trained with hard labels, even label smoothing improves teacher's accuracy}, as the authors found that label smoothing tends to ``erase'' information contained intra-class across individual examples, which indicates that the relative information between logits will be erased to some extent when the teacher is trained with label smoothing. This rising idea is becoming more and more dominant and has been quoted by a large number of recent literatures~\citep{arani2019improving,tang2020understanding,mghabbar2020building,shen2020mix,khosla2020supervised}.

\begin{figure}[t]
\centering
\includegraphics[width=0.99\linewidth]{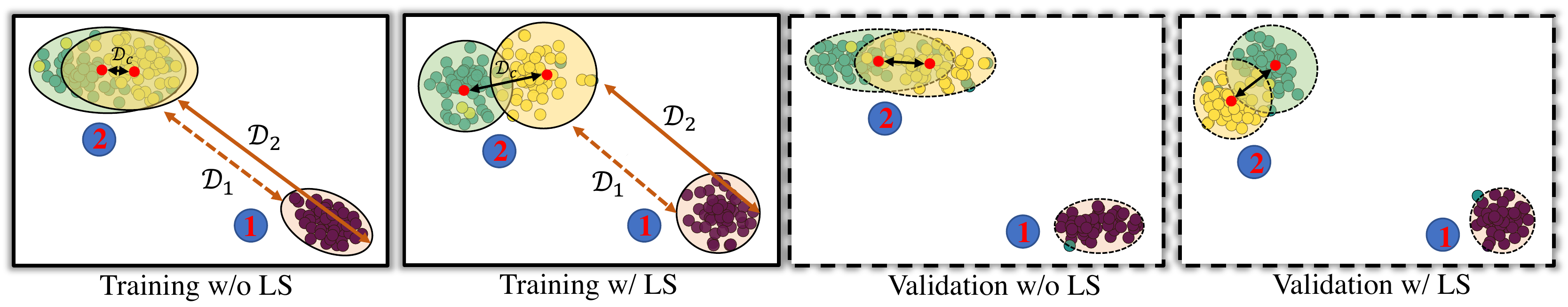}
\vspace{-0.12in}
\caption{Illustrations of the effects of label smoothing on penultimate layer output. The figure is plotted on ImageNet with ResNet-50 following~\citep{muller2019does}, we also choose two semantically similar classes (toy poodle and miniature poodle, in green and yellow) and one semantically different class (tench, in purple). \ding{172} is the discovery observed by M{\"u}ller et al. that label smoothing will enforce each example to be equidistant to its template, i.e., erasing the relative information between logits. $\mathcal{D}_1$ and $\mathcal{D}_2$ are the degree of measuring ``how much a tench is similar to poodle''. \ding{173} is our new finding in this paper that ``erasing'' effect enabled by label smoothing actually promotes to enlarge relative information on those semantically similar classes, i.e., making them have less overlap on representations. $\mathcal{D}_c$ is the distance between the semantically similar ``toy poodle'' cluster and the ``miniature poodle'' cluster. More details can be referred to Sec.~\ref{erase}.} 
\label{fig:enbedding}
\vspace{-0.35in}
\end{figure}

\vspace{-0.05in}
However, this seems reasonable observation basically has many inconsistencies in practice when adopting knowledge distillation with smoothing trained teachers. Thus, we would like to challenge whether this perspective is entirely correct? To make label smoothing and knowledge distillation less mysterious, in this paper, we first systematically introduce the mechanism and correlation between these two techniques. We then present a novel connection of label smoothing to the idea of ``erasing'' relative information. We expose the truth that factually {the negative effects of erasing relative information only happen on the semantically different classes.} Intuitively, those classes are easy to classify as they have obvious discrepancies. Therefore, the negative effects during distillation are fairly moderate. On those semantically similar classes, interestingly, we observe that {\bf erasing phenomenon can enforce two clusters being away from each other and actually enlarge the central distance of clusters between classes}, which means it makes the two categories easier for classifying, as shown in Fig.~\ref{fig:enbedding}. These classes in traditional training procedure are difficult to distinguish, so generally, the benefits of using label smoothing on teachers outweigh the disadvantages when training in knowledge distillation. Our observation in this paper supplements and consummates prior M{\"u}ller et al.'s discovery essentially, demonstrates that label smoothing is compatible with knowledge distillation through explaining the erasing logits information on similar classes. We further shed light on understanding the behavior and effects when label smoothing and knowledge distillation are applied simultaneously, making their connection more interpretable, practical and clearer for usage.

\vspace{-0.05in}
\noindent{\textbf{How to prove that their discovery is not judgmatic?}} 
We clarify such widely accepted idea through the following exploratory experiments, and exhaustively evaluate our proposed hypothesis: (i) Standard ImageNet-1K~\citep{deng2009imagenet}, fine-grained CUB200-2011~\citep{wah2011caltech} and noisy iMaterialist product recognition; (ii) Binary neural networks (BNNs)~\citep{rastegari2016xnor}; (iii) Neural machine translation (NMT)~\citep{vaswani2017attention}. Intriguingly, we observe that if the teacher is trained with label smoothing, the absolute values of converged distilling loss on training set are much larger than that the teacher is trained with hard labels, whereas, as we will discuss in detail later in Fig.~\ref{fig:cls_curves} and~\ref{fig:binary_curves}, the accuracy on validation set is still better than that without label smoothing.  
We explain this seemingly contradictory phenomenon through visualizing the teachers' output probabilities with and without label smoothing, it suggests that the suppression of label smoothing for knowledge distillation only happens on training phase as the distributions from teachers with label smoothing is more flattening, the generalization ability of networks on validation set is still learned during optimization. That is to say, the dynamical soft labels generated by teacher networks can prevent learning process from overfitting to the training data, meanwhile, improving the generalization on the unseen test data. Therefore, we consider this erasing relative information function within class from label smoothing as a merit to distinguish semantically similar classes for knowledge distillation, rather than a drawback. Moreover, we also propose a stability metric to evaluate the degree of erased information by label smoothing, we found the proposed metric is highly aligned with model's accuracy and can be regarded as a supplement or alternateness to identify good teachers for knowledge distillation. Finally, we discuss several intriguing properties of label smoothing we observed on the long-tailed category distribution and rapidly-increased \#class scenarios, as provided in Sec.~\ref{lost_eff}. \\More specifically, this paper aims to address the following questions: 

\vspace{-0.05in}
{\em $\bullet$ Does label smoothing in teacher networks suppress the effectiveness of knowledge distillation?}
Our answer is {\em No}. Label smoothing will not impair the predictive performance of students. Instead, we observe that a smoothing trained teacher can protect the student from overfitting on the training set, which means that with smoothing trained teachers in knowledge distillation, the training loss is always higher than that without smoothing, but the validation accuracy is still similar or even better. 

\vspace{-0.05in}
{\em $\bullet$ What will actually determine the performance of a student in knowledge distillation?}
From our empirical study, we observe if the student architecture is settled, the dominating factor in knowledge distillation is the quality of supervision, i.e., the performance of a teacher network. A higher-accuracy teacher is particularly successful in distilling a better student, regardless it is trained with or without label smoothing. 
This observation is partially against the conclusion in~\citep{muller2019does} which stated {``a teacher with better accuracy is not necessary to distill a better student''.}  

\vspace{-0.05in}
{\em $\bullet$ When will the label smoothing indeed lose its effectiveness for learning deep neural networks?} Long-tailed class distribution and increased number of classes are two scenarios we observed wherein label smoothing will lose or impair its effectiveness. We empirically verify the findings on iNaturalist 2019~\citep{van2018inaturalist}, Place-LT~\citep{liu2019large} and curated ImageNet~\citep{liu2019large}.

\begin{figure}[t]
\centering
\includegraphics[width=0.8\linewidth]{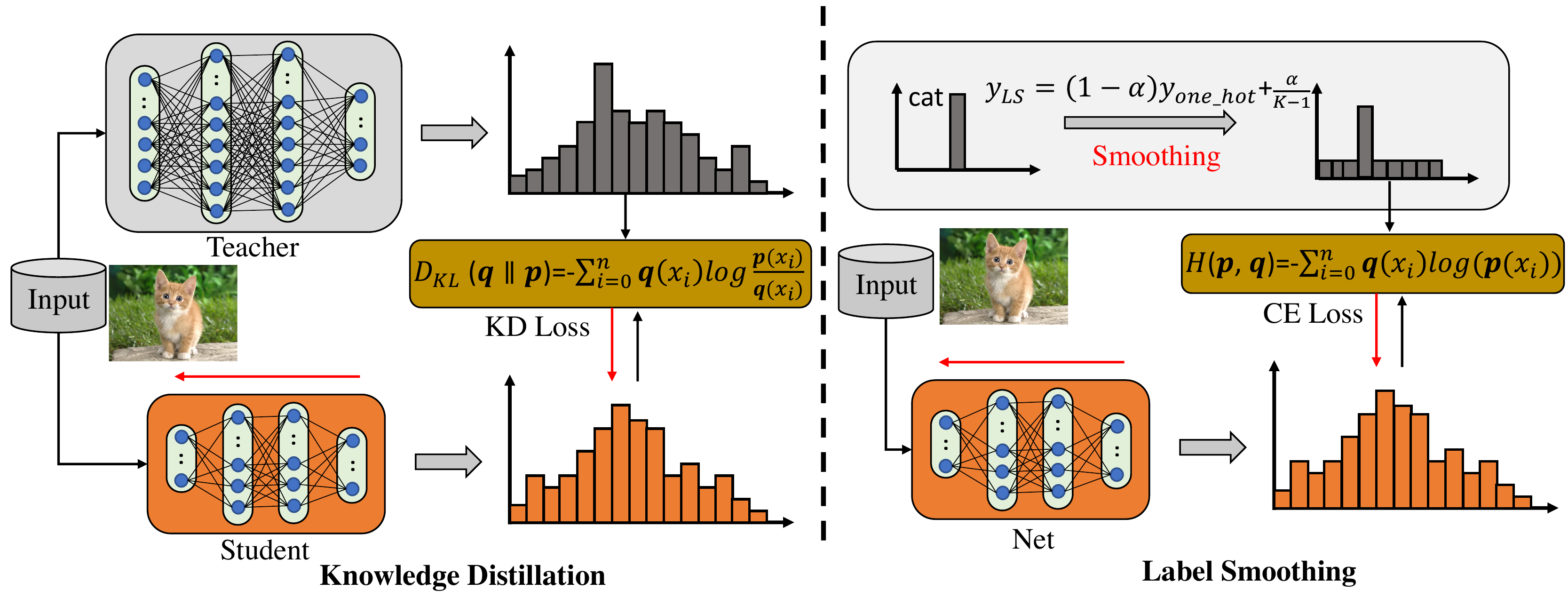}
\vspace{-0.15in}
\caption{Knowledge distillation (KD) and label smoothing (LS) overview. Both the KD and LS adopt softened distributions for learning the target networks. The KD differs from LS in the generation of these distributions and the objectives for optimization. KD chooses to utilize a pre-trained teacher to produce the supervision dynamically, while LS uses a constant uniform distribution for training. In the figure, the black lines are the forward pass and the red lines are the gradient propagation direction.}
\label{fig:comparison}
\vspace{-0.15in}
\end{figure}

\vspace{-0.05in}
\section{Background}
\vspace{-0.05in}
In this section, we first introduce the background of label smoothing and knowledge distillation through a mathematical description.
Given a dataset $\mathcal{D}={({X}, {Y})}$ over a set of classes $K$, ${X}$ is the input data and ${Y}$ is the corresponding one-hot label with each sample's label $\bm{y}\in \{0,1\}^K$, where the element $y_c$ is 1 for the ground-truth class and 0 for others. Label smoothing replaces one-hot hard label vector $\bm y$ with a mixture of weighted $\bm y$ and a uniform distribution:
\begin{equation}
y_{c}=\left\{\begin{array}{ll}1-\alpha & \text { if } c=label, \\ \alpha / (K-1) & \text { otherwise. }\end{array}\right.
\end{equation}
where $\alpha$ is a small constant coefficient for flattening the one-hot labels. Usually, label smoothing is adopted when the loss function is cross-entropy, and the network uses {\em softmax} function to the last layer's logits $\bm z$ to compute the output probabilities $\bm p$, so the gradient of each training sample with respect to $\bm z$ will be:
% \begin{equation}
  $ \nabla \mathcal{H}(\bm p, \bm y) = \bm p- \bm y=\sum\limits_{c=1}^K(\text{Softmax}(z_c)-y_c),$
% \end{equation}
where $\mathcal{H}(\bm p, \bm y)=-\sum\limits_{c=1}^Ky_c \text{log}p_c $ is the cross-entropy loss and $z_c$ is $c$-th logit in $\bm z$. \\

\begin{wrapfigure}{r}{6.5cm}
\vspace{-0.55in}
\centering
\includegraphics[width=0.99\linewidth]{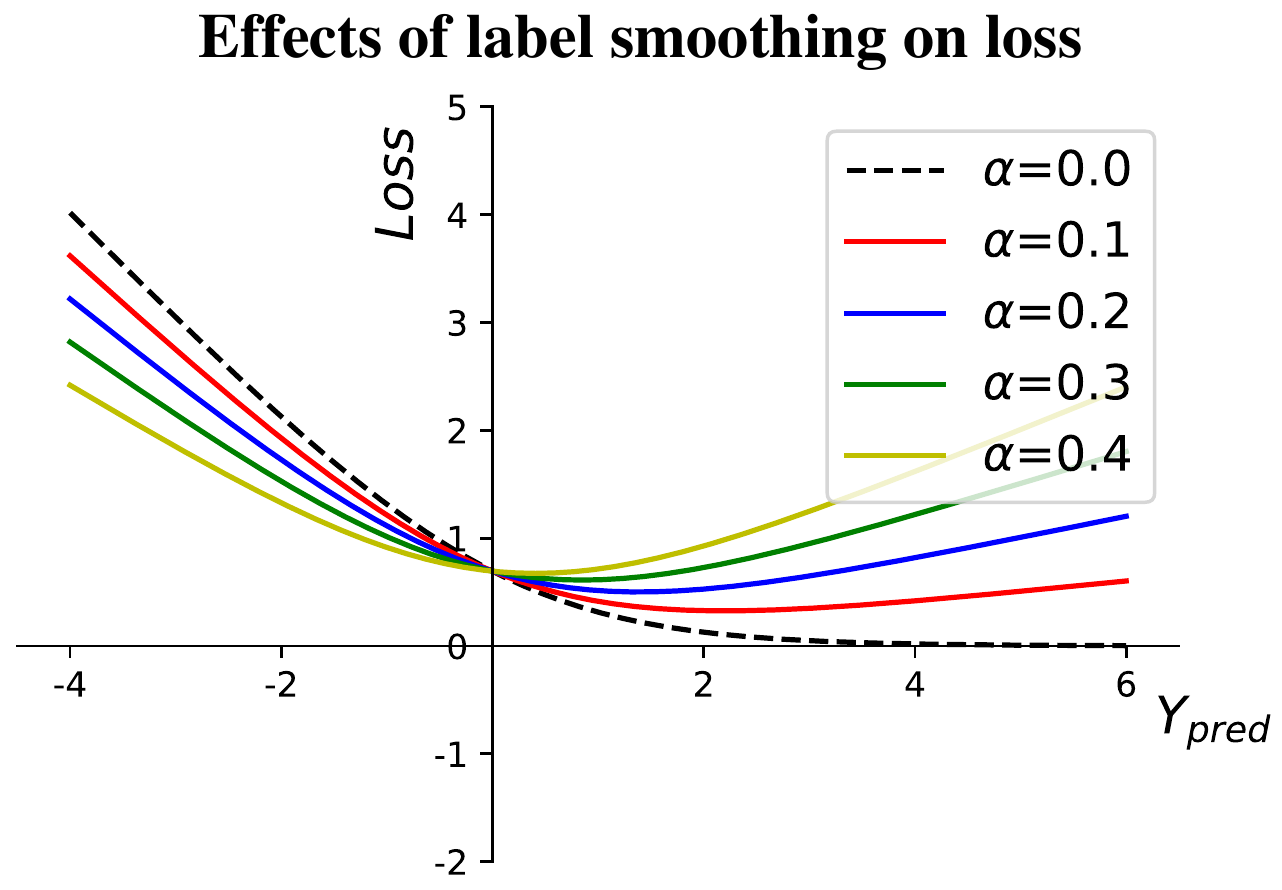}
\vspace{-0.3in}
\caption{Correction effects of label smoothing on logistic loss with different $\alpha$. Black dotted line presents the standard logistic loss and other colored lines are imposed label smoothing.}
\label{fig:ls_loss}
\vspace{-0.45in}
\end{wrapfigure}

To further understand the effects of label smoothing on loss function, Fig.~\ref{fig:ls_loss} illustrates correction effects of smoothing on the binary cross-entropy loss ($K=2$). We can observe that the standard logistic loss ($\alpha=0$) vanishes for large and confident positive predictions, and becomes linear for large negative predictions. Label smoothing will penalize confident predictions and involve a finite positive minimum as it aims to minimize the average per-class. Generally, larger $\alpha$ values will produce larger loss values rebounding at positive predictions. This is also the underlying reason that smoothed loss can flatten the predictions of a network.

In knowledge distillation, we usually pre-train the teacher model $\mathcal{T}_\mathbf{w}$ on the dataset in advance. The student model $\mathcal{S}_\mathbf{w}$ is trained over the same set of data, but utilizes labels generated by $\mathcal{T}_\mathbf{w}$. More specifically, we can regard this process as learning $\mathcal{S}_\mathbf{w}$ on a new labeled dataset $\tilde{\mathcal{D}}={(X, \mathcal{T}_\mathbf{w}(X))}$. Once the teacher network is trained, its parameters will be frozen in the whole distillation. 

The student network $\mathcal{S}_\mathbf{w}$ is trained by minimizing the similarity between its output and two parts: the hard one-hot labels and the soft labels generated by the teacher network. Letting $p_c^{{\mathcal T}_\mathbf{w}}({X})={\mathcal{T}_\mathbf{w} }({X})[c]$, $p_c^{{\mathcal S}_\mathbf{w}}({X})={\mathcal{S}_\mathbf{w} }({X})[c]$ be the probabilities assigned to class $c$ in the teacher model $\mathcal T_\mathbf{w}$ and student model $\mathcal S_\mathbf{w}$. The distillation loss can be formulated as $\
\lambda\mathcal{H}(\bm p^{{\mathcal S}_\mathbf{w}}, \bm y)+(1-\lambda)\mathcal{H}(\bm p^{{\mathcal S}_\mathbf{w}}/\mathcal{T}, \bm p^{\mathcal{T}_\mathbf{w}}/\mathcal{T})$ where $\mathcal{T}$ is the temperature scaling factor and $\lambda$ is the trade-off coefficient to balance the two terms.

\vspace{-0.05in}
\section{The ``Erase Information'' Effect by Label Smoothing} \label{erase}
\vspace{-0.05in}
This section aims to explain the erasing information effect more thoroughly. We start by reproducing the visualization of penultimate layer’s activations using the same procedure from~\citep{muller2019does}. We adopt ResNet-50 trained with hard and smoothed labels on ImageNet. As shown in Fig.~\ref{fig:enbedding}, we obtain similar distributions as~\citep{muller2019does}. Since examples in training set are the ones used for distillation, we mainly analyze the visualization from the training data. The core finding in~\citep{muller2019does} is that if a teacher is trained with hard labels, representations of examples are distributed in broad clusters, which means that different examples from the same class can have different similarities ($\mathcal{D}_1$ and $\mathcal{D}_2$) to other classes. For a teacher trained with label smoothing, they observed the opposite behavior. Label smoothing encourages examples to lie in tight equally separated clusters, so each example of one class has very similar proximities ($\mathcal{D}_1$ is closer to $\mathcal{D}_2$) to examples of the other classes. Our re-visualization also supports this discovery. 
The authors derive the conclusion that a teacher with better accuracy is not necessarily to distill a better student. This seems reasonable as the broad clusters can enable different examples from the same class to provide different similarities to other classes, which contains more information for knowledge distillation. 

However, if refocusing on the two semantically similar classes, when label smoothing is applied, the clusters are much tighter because label smoothing encourages each example is to be equidistant from all other class’s templates, while, the tight cluster substantially promotes different class representations to be separate, i.e., the distance of clusters $\mathcal{D}_c$ increases, which further indicates that different class examples obtain more distinguishable features. This phenomenon is crucial as these difficult classes are the key for boosting classification performance. Generally, it is not necessary to measure ``how much a poodle is a particularly similar to a tench'' since we have enough evidence to classify them, but it is critical to have information ``how different is a toy poodle to a miniature poodle''.

\begin{figure}[t]
\centering
\includegraphics[width=0.90\linewidth]{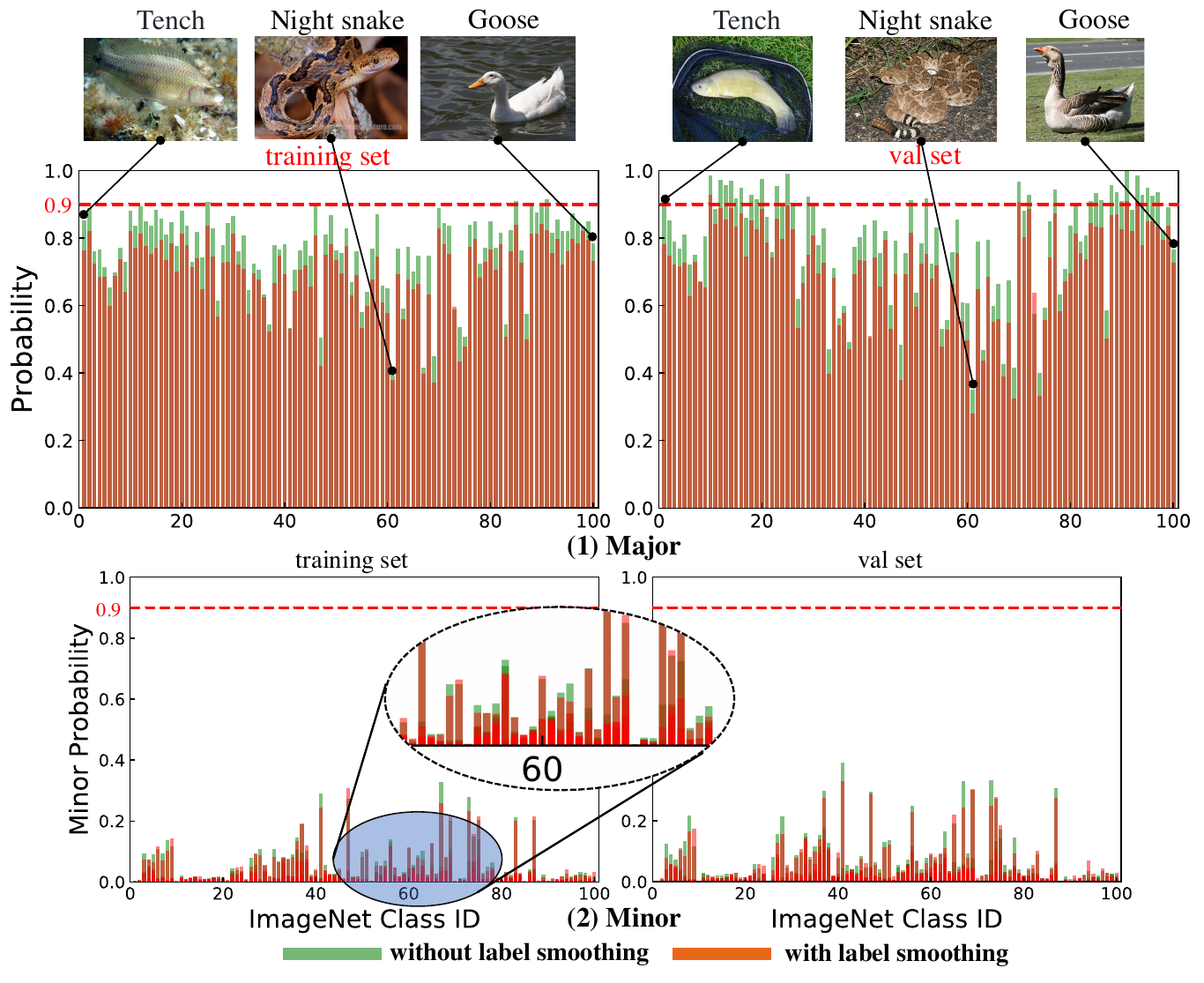}
\vspace{-0.15in}
\caption{Probability distributions with/without label smoothing on ResNet-50. We show the first 100 categories in ImageNet. The red/green bars are distributions with/without label smoothing, respectively. ``Minor probability'' denotes the small probabilities predicted by networks when the outputs are used as supervisions in knowledge distillation.}
\label{fig:weight_supervision}
\vspace{-0.2in}
\end{figure}

\noindent{\textbf{Visualizations of teacher predictions.}} We further visualization the mean distribution of different classes crossing examples, as shown in Fig.~\ref{fig:weight_supervision}. We average all the probabilities after softmax layer if the examples belong to the same category, and show the first 100 classes in ImageNet. Usually, the probabilities have a major value (the bars in Fig.~\ref{fig:weight_supervision} (1)) that represents model's prediction on category and other small values (i.e., minor predictions in Fig.~\ref{fig:weight_supervision} (2)) indicate that the input image is somewhat similar to those other categories, some discussions about minor predictions are given in Appendix~\ref{minor}. Our purpose of this visualization is to make certain of what label smoothing really calibrates in a network and shed light on how it affects the network predictions. We can observe in this figure that a model trained with label smoothing will generate more softened distributions, but the relations across different classes are still preserved. We conjecture the softened supervision is also the reason why teachers with label smoothing produce larger training loss during knowledge distillation. Consequently, {\em label smoothing will both decrease the variance (verified by following stability metric) and mean predictive values within a class, but will not impair the relations crossing different classes.}

\vspace{-0.05in}
\subsection{A Simple Metric for Measuring the Degree of Erased Information}
\vspace{-0.05in}
Different from the visualization scheme~\citep{muller2019does} of finding an orthonormal basis of the plane that only studies this problem qualitatively, we further address the ``erasing'' phenomenon through a statistical metric that is simple yet effective, and can measure the degree of erasing operation quantitatively. Our motivation behind it is straight-forward: If label smoothing erases relative information within a class, the variance of intra-class probabilities will decrease accordingly, thus we can use such variance to monitor the erasing degree, since this metric evaluates the fluctuation of the representations, we can also call it the stability metric. The definition is as follows:
\begin{equation}
\mathcal{S}_{\textrm{Stability}}=1-\frac{1}{K}\sum_{c=1}^{K}(\frac{1}{\bm n_c}\sum_{i=1}^{\bm n_c}||\bm p_{\{i,c\}}^{\mathcal T_\mathbf{w}}- \overline {\bm p}_{\{i,c\}}^{\mathcal T_\mathbf{w}}||^{2})
\end{equation}
where $i$ is the index of images and $\bm n_c$ is the \#image in class $c$. $\overline {\bm p}_{\{i,c\}}^{\mathcal T_\mathbf{w}}$ is the mean of $p^{\mathcal T_\mathbf{w}}$ in class $c$. This metric utilizes the probabilities of intra-class variance to measure the stability of a teacher’s prediction. The results on various network architectures are shown in Sec.~\ref{metric_eval} and a PyTorch-like code for calculating this metric is given in Appendix~\ref{alg}.

Such metric has at least two advantages: 
1) It can measure the degree of erased information quantitatively and further help discover more interesting phenomena, e.g., we observe that data augmentation method like CutMix~\citep{yun2019cutmix} together with longer training erases the relative information on logits dramatically and can further be reinforced by label smoothing.
2) We found that the proposed metric is highly aligned with model accuracy, thus such metric can be used as a complement for accuracy to evaluate the quality of teacher's supervision for knowledge distillation.

\vspace{-0.05in}
\section{A Close Look at Label Smoothing and Knowledge Distillation}
\vspace{-0.05in}

A few recent studies~\citep{shen2020meal,shen2019meal} suggested supervised part $\mathcal{H}(\bm p^{{\mathcal S}_\mathbf{w}}, \bm y)$ (i.e. hard labels) is not necessary as soft prediction is adequate to provide crucial information for students, meanwhile, removing supervised part can avoid involving incorrect labels caused by random crop, multi-object circumstance or false annotations by humans. Thus, here we only consider the soft part $\mathcal{H}(\bm p^{{\mathcal S}_\mathbf{w}}/\mathcal{T}, \bm p^{\mathcal{T}_\mathbf{w}}/\mathcal{T})$ with the commonly used {\em Kullback-Leibler divergence} similarity. \\
\noindent{\textbf{KL-divergence}} measures the similarity of two probability distributions. We train the student network $\mathcal{S}_\theta$ by minimizing the KL-divergence between its output $p_c^{{\mathcal S}_\theta}({X})$ and the soft labels $p_c^{{\mathcal T}_\theta}({X})$ generated by the teacher network. Following~\citep{muller2019does,hinton2015distilling} we set $\mathcal{T}=$1 as the temperature constant and it is omitted for simplicity, thus our loss function will be:
\begin{equation}
\begin{gathered}
\mathcal{D}_{K L}({\mathcal T}_\mathbf{w} \| {\mathcal S}_\mathbf{w})=\mathbb{E}_{x \sim {\mathcal T}_\mathbf{w}}\left[-\log \frac{{\mathcal S}_\mathbf{w}(X)}{{\mathcal T}_\mathbf{w}(X)}\right] \hfill \\
\quad \quad \quad  \quad \quad \quad \  =\mathbb{E}_{x \sim {\mathcal T}_\mathbf{w}}[-\log {\mathcal S}_\mathbf{w}(X)]-\mathcal{H}({\mathcal T}_\mathbf{w}(X))
\end{gathered} 
\end{equation}
Here, $\mathbb{E}_{x \sim {\mathcal T}_\mathbf{w}}[-\log {\mathcal S}_\mathbf{w}(X)]$ is the cross-entropy between ${\mathcal S}_\mathbf{w}$ and ${\mathcal T}_\mathbf{w}$ (denoted $\left.\mathcal{H}(\bm p^{{\mathcal S}_\mathbf{w}}, \bm p^{{\mathcal T}_\mathbf{w}})\right)$. The second term  $\mathcal{H}({\mathcal T}_\mathbf{w}(X))=\mathbb{E}_{x \sim {\mathcal T}_\mathbf{w}}[-\log \bm p^ {{\mathcal T}_\mathbf{w}}(x)]$ is the entropy of teacher ${\mathcal T}_\mathbf{w}$ and is constant with respect to $\mathcal{T}_\mathbf w$. We can remove it and simply minimize the  loss as follows:
\begin{equation} \label{kd_form}
\begin{gathered}
\mathcal{H}(\bm p^{{\mathcal S}_\mathbf{w}}, \bm p^{\mathcal{T}_\mathbf{w} })   =  - \sum\limits_{c=1}^K { {p_c^{{\mathcal{T}_\mathbf{w} }}({X})\log } } p_c^{{\mathcal{S}_\mathbf{w} }}({X}). \hfill  
\end{gathered} 
\end{equation}
We can observe that Eq.~\ref{kd_form} is actually a standard cross-entropy loss. Then, we have:

{\bf Property 1.} \ \ \  {\em If not consider hard labels in knowledge distillation, distillation loss and cross-entropy loss with label smoothing have the same optimizing objective, i.e., $\mathcal{D}_{K L}({\mathcal T}_\mathbf{w} \| {\mathcal S}_\mathbf{w})$ = $\mathcal{H}(\bm p^{{\mathcal S}_\mathbf{w}}, \bm p^{\mathcal{T}_\mathbf{w} })$.}

This property shows that label smoothing and knowledge distillation have the same optimization objective, the sole difference between them is the mechanism of producing the soft labels. Therefore, except for the neural machine translation, in this paper all of our knowledge distillation experiments are conducted without the hard labels, which means our student solely relies on the softened distribution from a teacher without the one-hot ground-truth. This may challenge the common practice in knowledge distillation~\citep{hinton2015distilling,romero2014fitnets} that  adopted both hard and soft labels with the cross-entropy loss for distillation, while our surprisingly good results and previous studies~\citep{shen2020meal,shen2019meal,bagherinezhad2018label} indicate that knowledge distillation enabled by soft labels solely is not only an auxiliary regularization~\citep{yuan2019revisit} but can be the dominating supervisions, which further inspires us to carefully revisit the role of knowledge distillation and design better supervision/objective in training deep neural networks. 

\vspace{-0.05in}
\section{Empirical Studies} \label{metric_eval}
\vspace{-0.05in}

\noindent{\textbf{Metric Evaluation.}}
Our results of stability metric are shown in Table~\ref{tab:stability-table}, the second and third columns are results without label smoothing and the last two are with it. We study the metric crossing a variety of different network architectures. The gaps of $\mathcal{S}_\textbf{Stability}$ using the same architecture measure the degree of erasing relative information. We can observe that the variances (1-$\mathcal{S}_\textbf{Stability}$) with label smoothing always have lower values than models trained without label smoothing, this proves that label smoothing will erase information and enforce intra-class representations of samples being similar. Generally, the accuracy and stability have a positive correlation between them. But the stability can even overcome some outliers, for example, Wide ResNet50 with label smoothing has lower accuracy, but the stability is still consistent with the tendency of predictive quality. Moreover, models trained with more epochs and augmentation techniques like CutMix~\citep{yun2019cutmix} can dramatically increase the stability, this means relative information will be erased significantly by more augmentation together with longer training. We emphasize that this discovery cannot be observed by the qualitative visualization method~\citep{muller2019does}. A PyTorch-like code is in Appendix~\ref{alg}.

\renewcommand{\arraystretch}{1.2}
\setlength{\tabcolsep}{0.5em}
\begin{table}[h]
\centering
\vspace{-0.1in}
\caption{Accuracy and stability results with and without label smoothing on ImageNet-1K. Here we show (1-$\mathcal{S}_\textbf{Stability}$), which denotes the aggregated intra-class variance (the lower the better). Red numbers are the quantitative values of the erased information by label smoothing.}
\label{tab:stability-table}
\vspace{-0.1in}
\resizebox{.99\textwidth}{!}{%
\begin{tabular}{lcc|cc}
 \toprule[1.1pt]
      \bf Netowrks             &\bf Acc. (\%) w/o LS     & \bf (1-$\mathcal{S}_\textbf{Stability}$) w/o LS  & \bf Acc. (\%) w/ LS    &\bf (1-$\mathcal{S}_\textbf{Stability}$) w/ LS       \\ \hline
{ResNet-18}~\citep{he2016deep}  & 69.758/89.078 & 0.3359 & \bf 69.774/89.122 & \bf 0.3358 (\color{red} -0.0001) \\ 
{ResNet-50}~\citep{he2016deep}  &75.888/92.642 &0.3217 & \bf 76.130/92.936 & \bf 0.3106 (\color{red}-0.0111) \\ 
{ResNet-101}~\citep{he2016deep} & 77.374/93.546 &0.3185 & \bf 77.726/93.830 & \bf 0.3070 (\color{red}-0.0115) \\ \hline
{MobileNet v2}~\citep{sandler2018mobilenetv2}  & 71.878/90.286& 0.3341 & -- & -- \\ \hline
{DenseNet121}~\citep{huang2017densely}  &  74.434/91.972 & 0.3243 & -- & -- \\ \hline
{ResNeXt50 32$\times$4d}~\citep{xie2017aggregated}  & 77.618/93.698&0.3229 & \bf 77.774/93.642 & \bf 0.3182 (\color{red}-0.0047) \\ \hline
{Wide ResNet50}~\citep{zagoruyko2016wide}  & \bf 78.468/94.086 &0.3201 &  77.808/93.682 & \bf 0.3155 (\color{red}-0.0046) \\ \hline
{ResNeXt101 32$\times$8d}~\citep{xie2017aggregated}  & 79.312/94.526 &0.3177 & \bf 79.698/94.768 & \bf 0.3116 (\color{red}-0.0061) \\ \hline
ResNet50+Long  &  76.526/93.070& 0.3222 &  \bf 77.106/93.340 & \bf 0.3090 (\color{red}-0.0132) \\ 
ResNet50+Long+CutMix~\citep{yun2019cutmix}  &  76.874/93.500 &  0.2999  & \bf 77.274/93.304 & \bf 0.2890 (\color{red}-0.0109) \\ 
\bottomrule[1.1pt]                 
\end{tabular}
}
\vspace{-0.1in}
\end{table}

\noindent{\textbf{Image Classification.}}
We verify our perspective through investigating the effectiveness of knowledge distillation with label smoothing on the image classification tasks. We conduct experiments on three datasets: ImageNet-1K~\citep{deng2009imagenet}, CUB200-2011~\citep{WahCUB_200_2011} and iMaterialist product recognition challenge data (in Appendix~\ref{detail_res}). We adopt ResNet-\{50/101\} as teacher networks and ResNet-\{18/50/101\} as students, respectively. More experimental settings are in Appendix~\ref{exp_setting}.

\begin{figure}[h]
\centering
\vspace{-0.15in}
\includegraphics[width=1\linewidth]{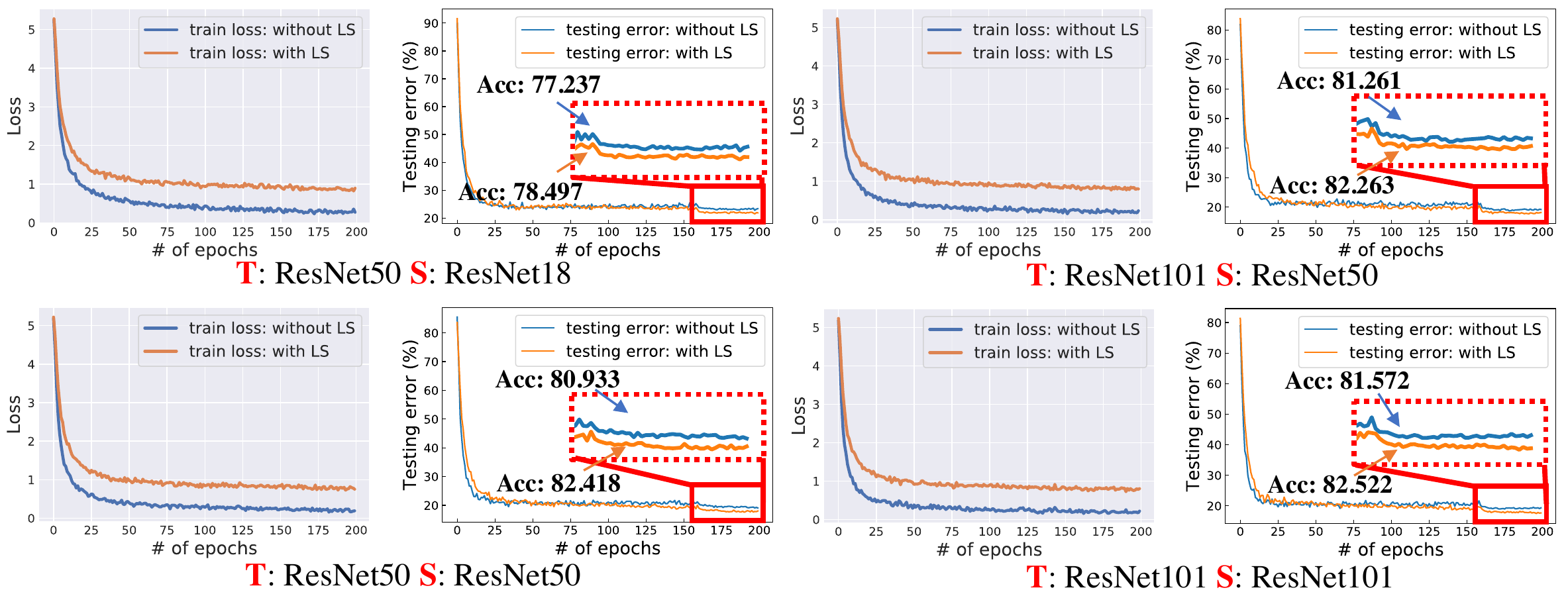}
\vspace{-0.28in}
\caption{The training and testing curves of knowledge distillation on CUB200-2011 when teachers are trained w/ and w/o label smoothing. The specific teacher and student architectures are given below each subfigure, therein, \textcolor{red}{T} indicates the teacher architecture and \textcolor{red}{S} indicates the student.}
\label{fig:cls_curves}
\vspace{-0.12in}
\end{figure}

\noindent{\textbf{Results.}}
The visualizations of our distillation training and testing curves are shown in Fig.~\ref{fig:cls_curves}. A more detailed comparison is listed in Tables~\ref{tab:cls-table} and~\ref{tab:my-table2}. From the visualization we found two interesting phenomena: On training set, the loss of teacher networks that trained with label smoothing is much higher than that of without label smoothing. While on validation set the accuracy is comparable or even slightly better (The boosts on CUB is greater than those on ImageNet-1K, as shown in Table~\ref{tab:cls-table}). We also provide the experiments of combining hard and soft labels in Appendix~\ref{hard}, the results still show the effectiveness of better teachers with label smoothing, which distill better students. To make it clearer why this happens in distillation, we visualize the supervisions from teacher networks in Fig.~\ref{fig:weight_supervision} and the discussion is shown there. It indicates that label smoothing flattens teacher's predictions which causes the enlarged training loss, while the student's generalization ability is still preserved.

\renewcommand{\arraystretch}{1.1}
\setlength{\tabcolsep}{1.1em}
\begin{table}[t]
\centering
\caption{Image classification results on ImageNet-1K, CUB200-2011 and iMaterialist product recognition (in Appendix~\ref{detail_res}). The teacher networks with label smoothing are denoted by ``\ding{52}''. We report average over 3 runs for all the teacher network training and student distillation.}
\label{tab:my-table}
\vspace{-0.15in}
\resizebox{0.9\textwidth}{!}{%
\begin{tabular}{ccccc}
\toprule[1.1pt]
\bf \em ImageNet-1K (Standard): \\ \hline
Teacher                     & w/ LS              & Acc. (Top1/Top5)               & Student    & Acc. (Top1/Top5) \\ \hline
    & \multirow{2}{*}{\ding{56}} & \multirow{2}{*}{76.056 $\pm$~0.119/92.791 $\pm$ 0.106} & ResNet-18  & 71.425 $\pm$~ 0.038/90.185 $\pm$ 0.075     \\
                            &                    &                                & ResNet-50  & 76.325 $\pm$~ 0.068/92.984 $\pm$ 0.043    \\ \cline{2-5} 
\rowcolor{mygray} \cellcolor{white} &   &                     & ResNet-18  & \bf 71.816 $\pm$~ \bf 0.017/90.466 $\pm$ 0.074     \\
  \rowcolor{mygray}\cellcolor{white} \multirow{-4}{*}{\large ResNet-50}  &  \multirow{-2}{*}{\ding{52}} &  \multirow{-2}{*}{\bf 76.128 $\pm$~\bf 0.069/\bf 92.977 $\pm$ \bf 0.030}  & ResNet-50  &  \bf 77.052 $\pm$~ \bf 0.030/93.376 $\pm$ 0.015   \\ %\hline 
\bottomrule[1.1pt]
\end{tabular}
}

\centering
\label{tab:cls-table}
\resizebox{0.9\textwidth}{!}{%
\begin{tabular}{ccccc}
% \toprule[1.5pt]
\bf \em CUB200-2011 (Fine-grained): \\ \hline
Teacher                     & w/ LS              & Acc. (Top1/Top5)               & Student    & Acc. (Top1/Top5) \\ \hline
 & \multirow{2}{*}{\ding{56}} & \multirow{2}{*}{79.931 $\pm$~0.037/94.370 $\pm$ 0.064} & ResNet-18  & 77.116 $\pm$~0.086/93.241 $\pm$ 0.108    \\
                            &                    &                                & ResNet-50  & 80.910 $\pm$~0.033/94.738 $\pm$ 0.114   \\ \cline{2-5} 
   \rowcolor{mygray} \cellcolor{white}                          &   &  & ResNet-18  &  \bf  78.382 $\pm$~\bf 0.099/93.621$\pm$ \bf 0.120     \\
 \rowcolor{mygray}\cellcolor{white} \multirow{-4}{*}{\large ResNet-50} &   \multirow{-2}{*}{\ding{52}}  &  \multirow{-2}{*}{\bf 81.497 $\pm$~\bf 0.035/95.043 $\pm$ 0.112}  & ResNet-50  &  \bf  82.355 $\pm$~\bf 0.050/95.440 $\pm$ \bf 0.075    \\ 
\bottomrule[1.1pt]
\vspace{-0.15in}
\end{tabular}
}
\vspace{-0.15in}
\end{table}

\noindent{\textbf{Binary Neural Networks (BNNs).}}
We then examine the effectiveness of knowledge distillation on Binary Neural Networks.
BNN aims to learn a network that both weights and activations are discrete values in \{-1, +1\}. In the forward pass, real-valued   activations are binarized by the sign function:
% \begin{align}
$\mathcal A_b = {\rm Sign}(\mathcal A_r) = \left\{  
             \begin{array}{lr}  
             - 1 & {\rm if} \ \ \mathcal A_r <0, \\  
             + 1 & {\rm otherwise.}  
             \end{array}  
\right. $
% \end{align}
where $\mathcal A_r$ is the real-valued activation of the previous layers, produced by the binary or real-valued convolution operations. $\mathcal A_b$ is the binarized activation. The real-valued weights are binarized by:
% \begin{align}
${\bf W}_b = \frac{||{\bf W}_r||_{l_1}}{n}  {\rm Sign}({\bf W}_r) = \left\{  
             \begin{array}{lr}  
             - \frac{|| {\bf W}_r||_{l_1}}{n}  & {\rm if} \ {\bf W}_r <0, \\ 
            %  \\
             + \frac{||{\bf W}_r||_{l_1}}{n}  & {\rm otherwise.}  
             \end{array} 
\right. $
% \end{align}
where ${\bf W}_r$ is the real-valued weights that are stored as {\em latent} parameters to accumulate the small gradients.  ${\bf W}_b$ is the binarized weights. We update binary weights through multiplying the sign of real-valued latent weights and the channel-wise absolute mean ($\frac{1}{n} ||{\bf W}_r||_{l_1}$).
Training BNNs is challenging as the gradient of optimization is approximated and the capacity of models is also limited.

\begin{figure}[h]
\vspace{-0.12in}
\centering
\includegraphics[width=0.65\linewidth]{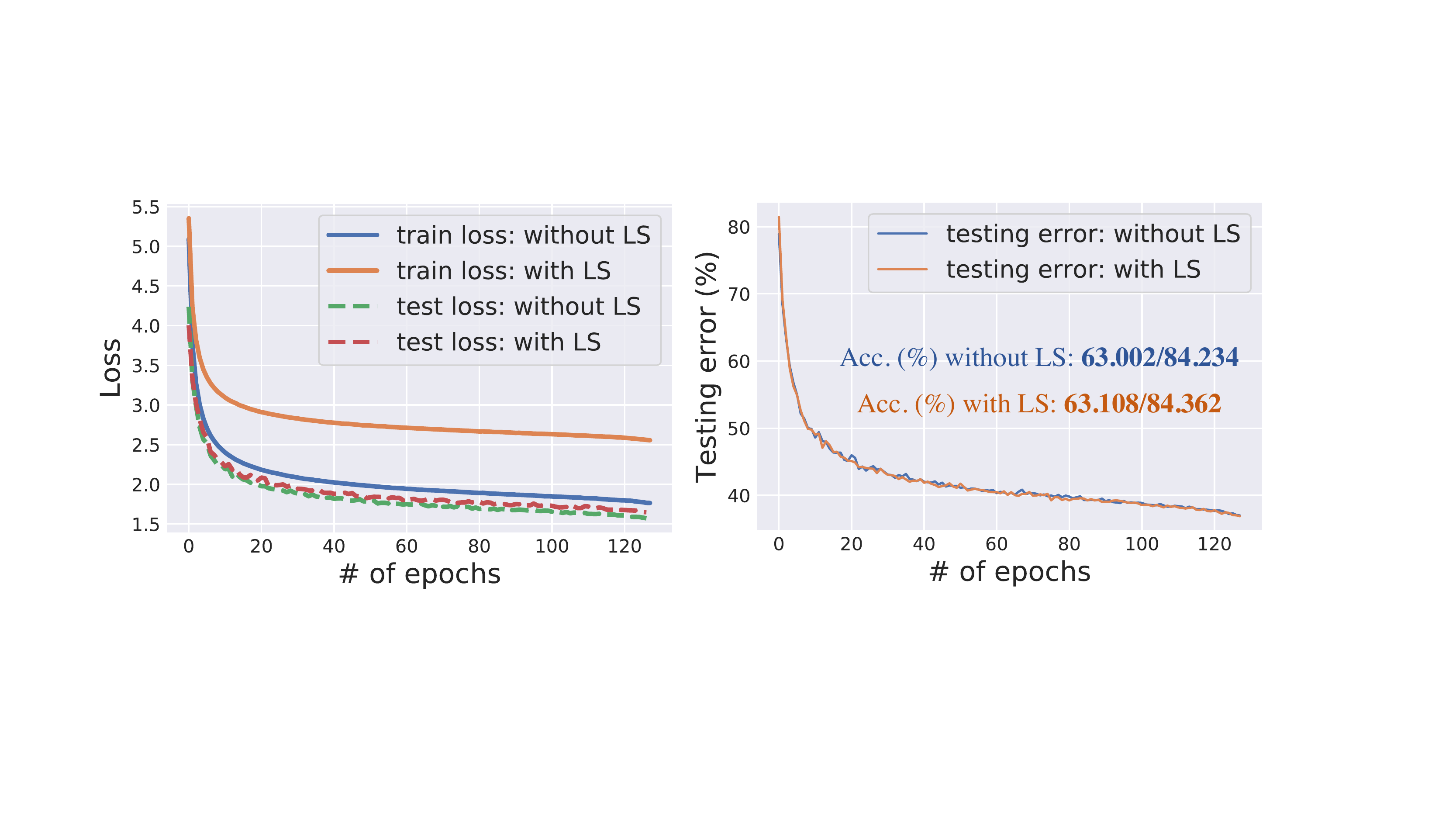}
\vspace{-0.1in}
\caption{Left is the averaged train/test loss curves in distillation, right is the testing error w/ best Top-1/5 accuracy. We use linear learning rate decay following other binary network training protocol~\citep{Martinez2020Training,liu2018bi}. Our teacher networks are ResNet-50 with and without label smoothing which have similar performance. The student is the state-of-the-art ReActNet~\citep{liu2020reactnet} with ResNet-18 backbone. We can observe that when the teacher is trained with label smoothing, the distillation loss is much higher, but the accuracy of student is still better.}
\label{fig:binary_curves}
\vspace{-0.05in}
\end{figure}

We perform experiments on ImageNet-1K and results are shown in Fig.~\ref{fig:binary_curves}. Withal, the teacher network trained with one-hot labels (blue curve) is over-confident as the loss value is much smaller, which means that the teacher trained with label smoothing can prevent distillation process from being over-confident on the training data, and obtain slightly better generalization and accuracy (63.108\% {\em vs.} 63.002\%) on the validation set. These results still support our conclusion on knowledge distillation.

\begin{wrapfigure}{r}{0.48\textwidth}
    \centering
    \vspace{-0.15in}
    \includegraphics[width=0.30\textheight]{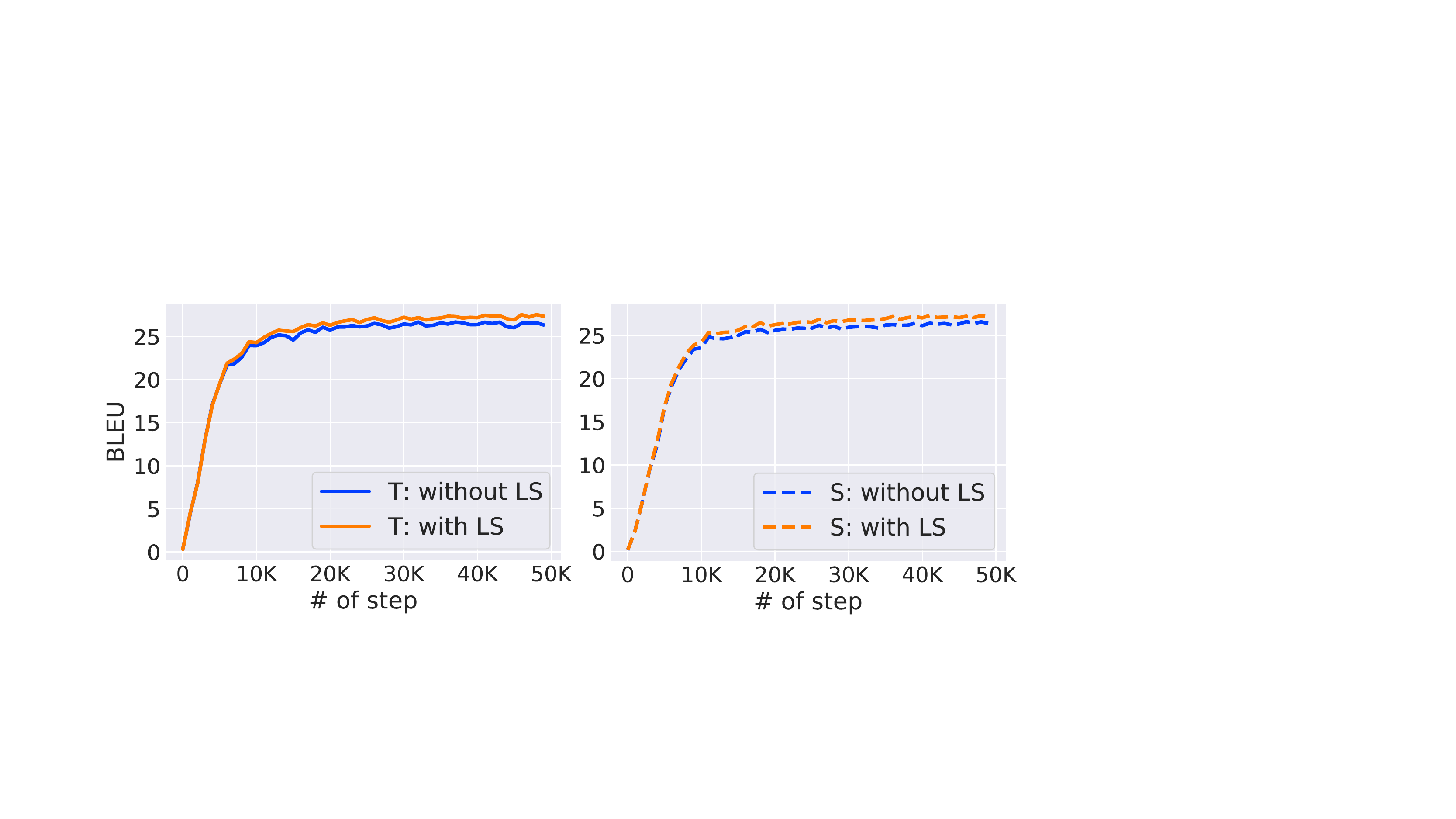}
    \vspace{-0.25in}
    \caption{Illustrations of BLEU score curves for teacher pre-training and student distillation. The left figure is teachers' pre-training with and without label smoothing. The right one is the distillation process of students.}
    \label{fig:nmt}
    \vspace{-0.15in}
\end{wrapfigure}
\vspace{-0.1in}
\noindent{\textbf{Neural Machine Translation (NMT).}} Finally, we investigate our hypothesis of knowledge distillation on the German-to-English translation task using the Transformer architecture~\citep{vaswani2017attention}. We utilize the distillation framework of ~\citep{tan2018multilingual} on IWSLT dataset, and the pre-training/distillation curves are shown in Fig.~\ref{fig:nmt}. A consistent setting is imposed on all the two comparison experiments, except the teacher is trained with and without label smoothing. 
We choose $\alpha=0.1$ for label smoothing as suggested by~\citep{vaswani2017attention,szegedy2016rethinking,muller2019does}, we use Adam~\citep{kingma2014adam} as the optimizer, {\em lr} with 0.0005, dropout with drop rate as 0.3, weight-decay with 0 and max tokens with 4096, all of these hyper-parameters are following the original settings of~\citep{tan2018multilingual}. 
Our results of Fig.~\ref{fig:nmt} deliver two important conclusions: First sub-illustration (left one) proves the statement of Vaswani et al.~\citep{vaswani2017attention} that label smoothing ($\alpha=0.1$) boosts the BLEU score of language model despite causing worse perplexity if comparing to a model is trained with one-hot/hard labels. Second sub-illustration (right one) implies that on the machine translation task, a stronger teacher (trained with label smoothing) will still distill a higher BLEU student. That is to say, label smoothing may not suppress the effectiveness of knowledge distillation in the NMT task.

\vspace{-0.05in}
\section{What Is A Better Teacher in Knowledge Distillation?}
\vspace{-0.05in}

\begin{figure}[h]
\vspace{-0.12in}
\centering
\includegraphics[width=0.74\linewidth]{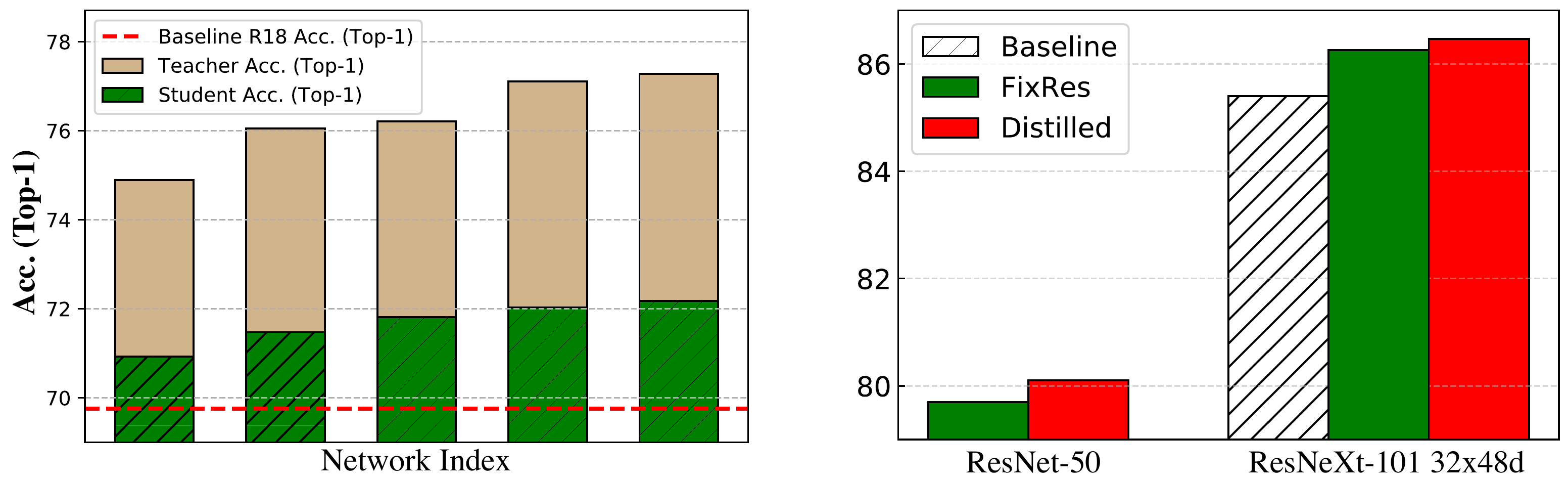}
\vspace{-0.1in}
\caption{Left is the accuracy relationship between teachers and students, wherein, all teachers are trained with label smoothing. Right is the accuracy of knowledge distillation by using strong teacher to fine-tune the student, FixRes~\citep{touvron2019FixRes} is adopted in both teacher and student networks.}
\label{fig:relation}
\vspace{-0.1in}
\end{figure}

\noindent{\bf Better Supervision is Crucial for Distillation.}
We further explore the effects of teacher's accuracy on the student through fixing the student structure and switching different teachers. We perform two settings for this ablation study: using the same teacher structure with different training strategies and different teacher architectures. All teacher models are re-trained with label smoothing. The results are shown in Fig.~\ref{fig:relation} (Left) and Table~\ref{tab:two_settings_table}, generally, teachers with higher accuracies can distill stronger students, but they are not linear related and are limited to the capability of the student itself. To further support the argument that {\em better teachers usually distill better students}, 
we choose the state-of-the-art FixRes model~\citep{touvron2019FixRes} for both the teacher and student and perform our distillation training via Eq.~\ref{kd_form}. The results are shown in Fig.~\ref{fig:relation} (Right) and our method is slightly better than the baseline and FixRes. Note that the compared FixRes is already the state-of-the-art with ResNet and ResNeXt architecture, so our result (under ResNet family) is a fairly competitive single-crop accuracy to date on ImageNet-1K.

\vspace{-0.06in}
\section{What Circumstances Indeed Will Make LS Less Effective?}\label{lost_eff}
\vspace{-0.07in}

\noindent{\textbf{Long-Tailed Distribution.}} We found the long-tailed datasets with imbalanced samples across classes will indeed suppress the effectiveness of label smoothing (LS) and hurt model performance. The results on long-tailed ImageNet-LT, Places365-LT and iNaturalist 2019 are shown in Table~\ref{tab:long-table_long_tail}.  
Our results on these three datasets empirically verify this observation and support this conclusion. Since LS has weight shrinkage and regularisation effects~\citep{lukasik2020does}, we have following conjecture:

{\bf Conjecture 1.} \ \ \  {\em Weight shrinkage effect (Regularisation) enabled by label smoothing is no longer effective on the long-tailed recognition circumstance and will further impair the performance.}

The weight shrinkage effect (label smoothing regularization) has been proven in~\citep{lukasik2020does} (Theorem 1) on the linear models. As label smoothing will assign probabilities equally to all minor classes (i.e., $(1-\alpha) \cdot \mathbf{I}_{pos}+\frac{\alpha}{L-1} \cdot \mathbf{J}_{neg}$), this operation may be biased to many-shot classes in the long-tailed scenario. We derive the conclusion empirically that label smoothing is inapplicable when the class distribution is long-tailed.

\begin{figure*}[h]
\vspace{-0.35in}
\begin{minipage}{0.65\textwidth}
    \centering\small
% \centering
\setlength{\tabcolsep}{2pt}
\captionof{table}{Teacher results on the curated ImageNet dataset when increasing the number of classes.}
\label{tab:increase-table}
\vspace{-0.1in}
\resizebox{.99\textwidth}{!}{%
\begin{tabular}{ccccc}
 \toprule[1.1pt]
                            &                    & ImageNet-100       & ImageNet-500         & ImageNet-1K         \\ \hline
Teacher                     & w/ LS              & Acc. (Top1/Top5)  & Acc. (Top1/Top5)  & Acc. (Top1/Top5)  \\
  &  {\ding{56}} & 82.380/\bf 95.520 &  73.521/91.642 & {\bf 69.758}/89.076 \\
                          
 \rowcolor{mygray} \cellcolor{white} \multirow{-2}{*}{ResNet-18}   & {\ding{52}} & {\bf 82.740}/95.440 & \bf 74.123/92.004 & 69.606/\bf 89.372 \\
                             \hline
 & {\ding{56}} & 82.000/94.340 & 81.712/95.080 & 77.374/93.546 \\
                            \rowcolor{mygray} \cellcolor{white} \multirow{-2}{*}{ResNet-101}   &  {\ding{52}} & \bf 83.400/95.300 & \bf 82.020/95.300 & \bf 77.836/93.662 \\ \hline
Average ($\uparrow$)                    &        &  \bf $\uparrow$0.880/0.440 & $\uparrow$0.455/0.291 &  $\uparrow$0.155/0.206 \\
\bottomrule[1.1pt]
\end{tabular}
}
\end{minipage}
 \hfill
\begin{minipage}{0.4\textwidth}
\vspace{18pt}
\centering\small
\centering
\includegraphics[width=0.68\linewidth]{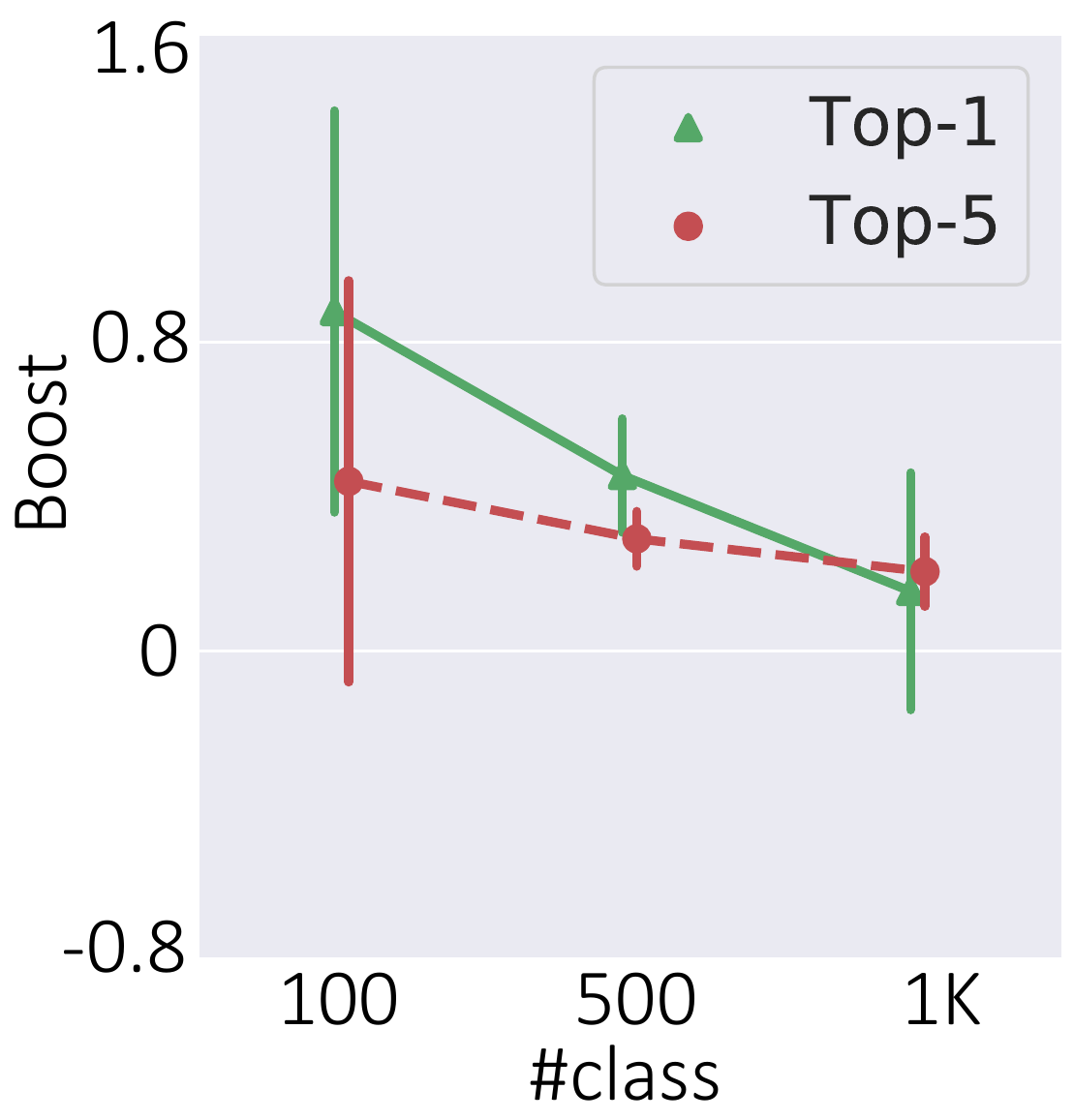}
\vspace{-0.15in}
\caption{Acc. downtrend.}
\label{tab:expr:combinatorial}
\end{minipage}

\end{figure*}

\noindent{\textbf{More \#Class.}} This is another circumstance we found will impair the effectiveness of label smoothing. The results are shown in Table~\ref{tab:increase-table} and Fig.~\ref{tab:expr:combinatorial}, we average the gains of label smoothing across two different network architectures and compare the boosts between curated ImageNet-100/500 and 1K. Generally, more classes will reduce the improvement produced by label smoothing.

\begin{table}[h]
\centering
\vspace{-0.1in}
\caption{Teacher results on the long-tailed ImageNet-LT, Places365-LT and iNaturalist 2019 val set.}
\label{tab:long-table_long_tail}
\vspace{-0.13in}
\resizebox{.75\textwidth}{!}{%
\begin{tabular}{ccccc}
 \toprule[1.1pt]
                            &                    & ImageNet-LT       & Place-LT          & iNaturalist 2019          \\ \hline
Teacher                     & w/ LS              & Acc. (Top1/Top5)  & Acc. (Top1/Top3)  & Acc. (Top1/Top3)  \\
\rowcolor{mygray} \cellcolor{white}  \multirow{2}{*}{ResNet-18}  &   {\ding{56}} &\bf 39.975/64.645 & \bf 26.479/47.233 & \bf 67.195/83.465 \\
                          
                            & {\ding{52}} & 39.115/63.655 & 25.877/46.260 & 66.700/83.432\\
                             \hline
\rowcolor{mygray} \cellcolor{white}   \multirow{2}{*}{ResNet-34}  &  {\ding{56}} &\bf 41.150/66.205 & \bf 27.329/48.753 & \bf 70.165/86.304 \\
                           
                            & {\ding{52}} & 40.965/65.850 & 26.863/48.110 &  69.406/86.073 \\
                             \hline
\rowcolor{mygray} \cellcolor{white}   \multirow{2}{*}{ResNet-50}  &  {\ding{56}} &\bf 40.985/66.030 & 27.384/\bf 48.740 & \bf 73.729/88.845 \\
                           
                            & {\ding{52}} & 39.965/65.195 & {\bf 27.562}/47.945 & 72.904/87.954 \\
                             \hline
\rowcolor{mygray} \cellcolor{white}  \multirow{2}{*}{ResNet-101} & {\ding{56}} & --/-- & \bf 28.096/50.164 & {\bf 74.389}/88.416 \\
                            
                            & {\ding{52}} & --/-- & 27.466/48.781 & 73.597/\bf 88.779 \\
\bottomrule[1.1pt]                 
\end{tabular}
}
\vspace{-0.1in}
\end{table}

\vspace{-0.35in}
\section{Conclusion}
\vspace{-0.06in}
We empirically demonstrated that label smoothing could both decrease the variance (i.e., erase relative information between logits) and lower mean predictive values (i.e., make prediction less confident) within a category, but it does not impair the relation distribution across different categories. Our results on image classification, binary neural networks, and neural machine translation indicate that label smoothing is compatible with knowledge distillation and this finding encourages more careful to understand and utilize the relationships of label smoothing and knowledge distillation in practice. We found through extensive experiments and analyses that the indeed circumstances label smoothing will lose its effectiveness are long-tailed distribution and increased number of classes. Our study also suggests that, to find a better teacher for knowledge distillation, accuracy of teacher network is one factor, the stability of supervision from teacher network is also an alternative indicator.

\bibliography{iclr2021_conference}
\bibliographystyle{iclr2021_conference}

\newpage

\appendix
\section*{\Large{Appendix}}
\vspace{-1ex}

In this appendix, we provide details omitted in the main text, including:

• Section~\ref{exp_setting}: A introduction of datasets and experimental settings (Sec. 5 ``Empirical Studies'' of the main paper.)

• Section~\ref{inferior}: A discussion about what happens if teachers with label smoothing are inferior? (Sec. 6 ``What Is a Better Teacher in Knowledge Distillation?''
of the main paper.)

• Section~\ref{alg}: A PyTorch-like code for calculating stability metric (Sec. 3.1 “A Simple Metric” of the main paper.)

• Section~\ref{detail_res}: (1) Table~\ref{tab:my-table2} is a detailed version of our Table~\ref{tab:my-table} (Sec. 5 ``Empirical Studies'' of the main paper.) 
(2) Table~\ref{tab:two_settings_table} is a detailed results for visualization of Fig.~\ref{fig:relation} (Sec. 6 ``What Is a Better Teacher in Knowledge Distillation?'' of the main paper.)

• Section~\ref{hard}: Results of knowledge distillation by adopting both hard and soft labels (Sec. 5 ``Empirical Studies'' of the main paper.)

• Section~\ref{minor}: Discussions of minor probabilities from teacher network. (Sec. 3 ``The Erase Information Effect by Label Smoothing'' of the main paper.)

• Section~\ref{inter}: Supplementary metric of inter-class variation. (Sec. 5 ``Metric Evaluation'' of the main paper.)

\section{Datasets and Experimental Settings} \label{exp_setting}

\subsection{For Training Teacher and Knowledge Distillation Models to Verify the Compatibility between Label Smoothing and Knowledge Distillation}
\noindent{\textbf{Standard ImageNet-1K Classification~\citep{deng2009imagenet}} ImageNet-1K contains $\sim$1.28 million images in 1000 classes for training and 50K images for validation. For training teacher networks, we follow the standard training protocol~\citep{he2016deep,goyal2017accurate}, i.e., total training epoch is 90, initial learning rate is 0.1 and decayed to 1/10 with every 30 epochs. For distillation, as the supervision is a soft distribution and will dynamically change, we train with 200 epochs and the learning rate is multiplied by 0.1 at 80 and 160 epochs. All models are trained from scratch. 

\noindent{\textbf{Fine-grained Recognition on CUB200-2011~\citep{WahCUB_200_2011}} This dataset contains 200 bird species and 11,788 images, and is a wildly-used fine-grained classification benchmark. Following~\citep{WahCUB_200_2011}, we use standard split with $\sim$30 samples of each breed for both training and testing. As CUB200-2011 has limited training data, we fine-tune our model from the ImageNet pre-trained parameters. Both teachers and students are trained with 200 epochs and the learning rate is multiplied by 0.1 at 80 and 160 epochs.

\noindent{\textbf{Noisy Product Recognition (iMaterialist Challenge)\footnote{\url{https://sites.google.com/view/fgvc6/competitions/imat-product-2019?authuser=0}.}.}} The training data consists of 1,011,532 images from 2,019 categories. The validation data has 10,095 images (around 5 for each category). The training data is collected from the Internet and contains some noise ($\sim$30\% of incorrect labels), the validation data has been cleaned by human annotators. The training protocol is following CUB200-2011. While, as the training images in this dataset is sufficiently enough, all our models are trained from scratch. 

\subsection{For Training Teachers Only to Explore the Effectiveness of Label Smoothing}
\noindent{\textbf{Long-tailed iNaturalist 2019~\citep{van2018inaturalist}.}} The iNaturalist Challenge 2019 dataset contains 1,010 species, with a combined training and validation set of 268,243 images, wherein, the validation set has a uniform distribution of three images in each category. While, the training set is constructed with a long-tailed distribution of image numbers. We conduct experiments using the same training protocol as the CUB200-2011 dataset. we evaluate our results on validation set using Top-1/Top-3 metrics. Also, this protocol is applied to the Place-LT dataset.

\noindent{\textbf{Place-LT~\citep{liu2019large}.}} The training set of this dataset is constructed by a Pareto distribution~\citep{William2001} with a power value $\alpha=6$. It contains a total number of 184.5K images from 365 categories. There are 4,980$\sim$5 images per class from the 365 classes of Places-365 dataset~\citep{zhou2017places}. 20 images per class are randomly selected from the original Places-365 training set as a validation set. The training protocol is the same as iNaturalist 2019, we  also evaluate our results on validation set using Top-1/Top-3 metrics. 

\noindent{\textbf{ImageNet-LT~\citep{liu2019large}.}} The long-tailed ImageNet is constructed by sampling a subset from the standard ImageNet~\citep{deng2009imagenet} following the Pareto distribution with the power value $\alpha=6$. The whole dataset contains 115.8K images from 1000 categories, with maximal 1280 images and minimal 5 images per class. 20 training images per class are randomly selected from the origin training set of ImageNet as a validation set. All our models are trained from scratch.

\noindent{\textbf{Curated ImageNet-100/500/1K.}} We construct this dataset by randomly selecting a subset of 100/500/1K categories in the standard ImageNet~\citep{deng2009imagenet}. We use this dataset to explore the impact of \#class for label smoothing. We observe that label smoothing is more effective on the dataset with fewer numbers of classes.

\renewcommand{\arraystretch}{1.0}
\setlength{\tabcolsep}{0.2em}
\begin{table}[h]
\centering
\caption{Overview of six datasets used in our experiments. $\alpha$ indicates the Pareto distribution value.}
\label{tab:my-table_data}
\vspace{-0.15in}
\resizebox{1.0\textwidth}{!}{%
\begin{tabular}{l|c|c|c|c|c|c|c}
\toprule[1.5pt]
\bf Dataset & \#class & Property         & Long-tailed $\alpha$ & Head class size & Tail class size & \#Training set & \#Testing set \\ \midrule \midrule
ImageNet-1K    &  1000    & Standard     & N/A &  N/A     & N/A & 1,281,167 & 50,000 \\  
CUB200-2011    &  200     & Fine-grained & N/A & N/A     &  N/A & 5,994 & 5,794 \\
iMaterialist   &  2019     & Noisy ($\sim$30\%) & N/A&  N/A  & N/A  & $\sim$1M     & $\sim$10K \\ \midrule
iNaturalist 2019 & 1010  & Long-tailed   & N/A  & $\sim$250 & $\sim$210 & 265,213 & 3,030  \\
ImageNet-LT &       1000 & Long-tailed   &  6  & 389 &  611     & 115,846     & 20,000\\
Place-LT &         365 & Long-tailed     &  6  & $\sim$100 &  $\sim$100  & 184,500  & 7,300\\
\bottomrule[1.5pt]
\end{tabular}
}
% \vspace{-0.15in}
\end{table}

\section{What happens if teachers with label smoothing are inferior?}\label{inferior}

In the main paper, we discussed that the superior teachers (even trained with label smoothing) could distill better students. 
Also, as we noted that long-tailed distribution will suppress the effectiveness of label smoothing. In this section, we would like to explore what happens if we use these inferior teachers for knowledge distillation? We conduct experiments on the long-tailed iNaturalist 2019 data and show more evidence to prove our conclusion that the quality of supervision is more crucial for knowledge distillation than the ways of training teachers (with or without label smoothing). The results are shown in Table~\ref{tab:long-table_bad}, it can be observed that if teachers with lower accuracies, the distilled students will also have poorer performance, regardless of the teacher is trained with label smoothing or not.

\begin{table}[h]
\centering
\caption{Distillation results using inferior teachers on the long-tailed iNaturalist 2019~\citep{van2018inaturalist} .}
\label{tab:long-table_bad}
\vspace{-0.1in}
\resizebox{.83\textwidth}{!}{%
\begin{tabular}{ccccc}
\toprule[1.5pt]
\em iNaturalist 2019 (Long-tailed) \\ \hline
Teacher                     & w/ LS              & Acc. (Top1/Top3)               & Student    & Acc. (Top1/Top3) \\ \hline
\multirow{4}{*}{ResNet-50}  & \multirow{2}{*}{\ding{56}} & \multirow{2}{*}{\bf 73.729/88.845} & ResNet-18  &\bf 67.756/84.521  \\
                            &                    &                                & ResNet-50  & \bf 74.125/88.944 \\ \cline{2-5} 
                            & \multirow{2}{*}{\ding{52}} & \multirow{2}{*}{72.904/87.954} & ResNet-18  &    67.228/84.587              \\
                            &                    &                                & ResNet-50  &    72.838/88.152    \\ \hline 
\multirow{4}{*}{ResNet-101} & \multirow{2}{*}{\ding{56}} & \multirow{2}{*}{\bf 74.389/88.416} & ResNet-50  & \bf 73.894/89.142     \\
                            &                    &                                & ResNet-101 & \bf  74.488/89.109    \\ \cline{2-5} 
                            & \multirow{2}{*}{\ding{52}} & \multirow{2}{*}{73.597/88.779} & ResNet-50  &   72.409/87.657         \\
                            &                    &                                & ResNet-101 &  73.234/88.647  \\
\bottomrule[1.5pt]
\end{tabular}
}
\end{table}

% \newpage
\section{A PyTorch-like Code for Calculating Stability Metric} \label{alg}
% ##################################################################################################
\begin{algorithm}[h]
\caption{PyTorch-like Code for Calculating Stability Metric.}
\label{alg:code}
\definecolor{codeblue}{rgb}{0.25,0.5,0.5}
\lstset{
  backgroundcolor=\color{white},
  basicstyle=\fontsize{7.2pt}{7.2pt}\ttfamily\selectfont,
  columns=fullflexible,
  breaklines=true,
  captionpos=b,
  commentstyle=\fontsize{7.2pt}{7.2pt}\color{codeblue},
  keywordstyle=\fontsize{7.2pt}{7.2pt},
  moredelim=**[is][\color{red}]{@}{@},
}
\lstdefinestyle{base}{
  moredelim=**[is][\color{red}]{@}{@},
}
\begin{lstlisting}[language=python,,style=base]
# x, target: input images and the corresponding labels
# single_std: std value for each class
# N: batchsize (we use N=50 on ImageNet for ease of implementation as it has 50 val images in each class)
# model_path: model path that you want to calculate stability

# load model
checkpoint = torch.load(model_path)
model.load_state_dict(checkpoint['state_dict']) # initialize
for (x, target) in loader:  # load a minibatch x with N samples (we choose N=50 for ImageNet)
    # make sure shuffle = False in dataloader

    # compute output
    output = model(x)   

    # compute softmax probabilities
    softmax_p = nn.functional.softmax(output, dim=1) 
    predict = softmax_p[:, target[0]]   # obtain target probability
    
    # compute stability
    single_std = predict.std(dim=0)
    all_std += single_std   # aggregate all std values from each class
    
return 1 - all_std/1000.0   # return computed stability
\end{lstlisting}
\end{algorithm}
% ##################################################################################################

\newpage
\section{Two Detailed Tables of Results} \label{detail_res}

\renewcommand{\arraystretch}{1.2}
\setlength{\tabcolsep}{1.1em}
\begin{table}[h]
\centering
\caption{Image classification results on ImageNet-1K, CUB200-2011 and iMaterialist product recognition. The teacher networks with label smoothing are denoted by ``\ding{52}''. We report average over 3 runs for ResNet-50 as the teacher network on both teacher training and knowledge distillation.}
\label{tab:my-table2}
\vspace{-0.15in}
\resizebox{1.0\textwidth}{!}{%
\begin{tabular}{ccccc}
\toprule[1.1pt]
\bf \em ImageNet-1K (Standard): \\ \hline
Teacher                     & w/ LS              & Acc. (Top1/Top5)               & Student    & Acc. (Top1/Top5) \\ \hline
  & \multirow{2}{*}{\ding{56}} & \multirow{2}{*}{76.056 $\pm$~0.119/92.791 $\pm$ 0.106} & ResNet-18  & 71.425 $\pm$~ 0.038/90.185 $\pm$ 0.075     \\
                            &                    &                                & ResNet-50  & 76.325 $\pm$~ 0.068/92.984 $\pm$ 0.043    \\ \cline{2-5} 
    \rowcolor{mygray} \cellcolor{white}      &    &                           & ResNet-18  & \bf 71.816 $\pm$~ \bf 0.017/90.466 $\pm$ 0.074     \\
     \rowcolor{mygray} \cellcolor{white}  \multirow{-4}{*}{\large ResNet-50}   &  \multirow{-2}{*}{\ding{52}} &  \multirow{-2}{*}{\bf 76.128 $\pm$~\bf 0.069/\bf 92.977 $\pm$ \bf 0.030}  & ResNet-50  &  \bf 77.052 $\pm$~ \bf 0.030/93.376 $\pm$ 0.015   \\ \hline 
& \multirow{2}{*}{\ding{56}} & \multirow{2}{*}{77.374/93.546} & ResNet-50  & 77.428/93.712    \\
                            &                    &                                & ResNet-101 & 78.270/94.152    \\ \cline{2-5} 
     \rowcolor{mygray} \cellcolor{white}        &  &                            & ResNet-50  & \bf 77.624/93.862    \\
     \rowcolor{mygray} \cellcolor{white}  \multirow{-4}{*}{\large ResNet-101}   &   \multirow{-2}{*}{\ding{52}}  &  \multirow{-2}{*}{\bf 77.836/93.662}    & ResNet-101 &  \bf 78.476/94.008   \\
\bottomrule[1.1pt]
\end{tabular}
}

\centering
\label{tab:cls-table2}
\resizebox{1.0\textwidth}{!}{%
\begin{tabular}{ccccc}
% \toprule[1.5pt]
\bf \em CUB200-2011 (Fine-grained): \\ \hline
Teacher                     & w/ LS              & Acc. (Top1/Top5)               & Student    & Acc. (Top1/Top5) \\ \hline
 & \multirow{2}{*}{\ding{56}} & \multirow{2}{*}{79.931 $\pm$~0.037/94.370 $\pm$ 0.064} & ResNet-18  & 77.116 $\pm$~0.086/93.241 $\pm$ 0.108    \\
                            &                    &                                & ResNet-50  & 80.910 $\pm$~0.033/94.738 $\pm$ 0.114   \\ \cline{2-5} 
   \rowcolor{mygray} \cellcolor{white}     &   &  & ResNet-18  &  \bf  78.382 $\pm$~\bf 0.099/93.621$\pm$ \bf 0.120     \\
    \rowcolor{mygray} \cellcolor{white}  \multirow{-4}{*}{\large ResNet-50}   &   \multirow{-2}{*}{\ding{52}}  &  \multirow{-2}{*}{\bf 81.497 $\pm$~\bf 0.035/95.043 $\pm$ 0.112}  & ResNet-50  &  \bf  82.355 $\pm$~\bf 0.050/95.440 $\pm$ \bf 0.075    \\ \hline 
& \multirow{2}{*}{\ding{56}} & \multirow{2}{*}{80.380/94.491} & ResNet-50  & 81.261/94.905    \\
                            &                    &                                & ResNet-101 &   81.572/\bf 95.371   \\ \cline{2-5} 
      \rowcolor{mygray} \cellcolor{white}     &   & & ResNet-50  & \bf  82.263/95.320    \\
    \rowcolor{mygray} \cellcolor{white}    \multirow{-4}{*}{\large ResNet-101}     &   \multirow{-2}{*}{\ding{52}} & \multirow{-2}{*}{\bf 82.332/94.970}   & ResNet-101 & {\bf 82.522}/95.199  \\ 
\bottomrule[1.1pt]
\end{tabular}
}

\centering
\label{tab:my-table2}
\resizebox{.83\textwidth}{!}{%
\begin{tabular}{ccccc}
\bf \em  iMaterialist-2019\_P (Noisy): \\ \hline
Teacher                     & w/ LS              & Acc. (Top1/Top3)               & Student    & Acc. (Top1/Top3) \\ \hline
  & \multirow{2}{*}{\ding{56}} & \multirow{2}{*}{66.241/91.015} & ResNet-18  &    65.250/90.243  \\
                            &      &                                & ResNet-50  &  67.420/92.155   \\ \cline{2-5} 
  \rowcolor{mygray} \cellcolor{white}     &  &           & ResNet-18  & \bf 65.359/90.530        \\
   \rowcolor{mygray} \cellcolor{white}  \multirow{-4}{*}{\large ResNet-50}  &   \multirow{-2}{*}{\ding{52}}  &  \multirow{-2}{*}{\bf  66.825/91.669  }  & ResNet-50  &  \bf  67.528/92.551   \\ \hline 
 & \multirow{2}{*}{\ding{56}} & \multirow{2}{*}{66.726/91.263} & ResNet-50  &  67.905/92.481   \\
                            &                    &                                & ResNet-101 &   68.281/92.580   \\ \cline{2-5} 
  \rowcolor{mygray} \cellcolor{white}    &      &                               & ResNet-50  & \bf 67.925/92.789     \\
 \rowcolor{mygray} \cellcolor{white}   \multirow{-4}{*}{\large ResNet-101}    &    \multirow{-2}{*}{\ding{52}}   &  \multirow{-2}{*}{\bf 67.370/91.877}  & ResNet-101 & \bf 68.618/92.907  \\ 
\bottomrule[1.1pt]
\end{tabular}
}
\end{table}

{
\renewcommand{\arraystretch}{1.0}
\setlength{\tabcolsep}{0.2em}
\begin{table}[h]
\centering
\caption{ImageNet results on the same student structure with different teachers. ``Long'' indicates we train with more budget (160 epochs), the default is 90. ``R50 and R18'' are ResNet-50/18, respectively.}
\label{tab:two_settings_table}
\vspace{-0.13in}
\resizebox{1.0\textwidth}{!}{%
\begin{tabular}{lccc|lccc}
\toprule[1.5pt]
\bf Teacher (same arch) &\bf Acc. (Top-1) & \bf Student &\bf Acc. (Top-1) & \bf Teacher (different archs)   & \bf Acc. (Top-1) & \bf Student & \bf Acc. (Top-1) \\
R50                 &  76.056    & R18     &   71.478 & MobileNet V2 & 71.878 & R18 &  70.054\\  
R50+LS              & 76.212 & R18     & 71.816 & DenseNet-121  &  74.894  & R18     & 70.922 \\
R50+LS+Long        & 77.106& R18     & 72.024  & Wide ResNet-50-2   &  77.808   & R18     & 72.232  \\
R50+LS+Long+CutMix & 77.274 & R18     & 72.172  & ResNeXt-101-32x8d &  79.698   & R18     & 72.412\\ 
\bottomrule[1.1pt] 
\end{tabular}
}
\end{table}
}

\begin{table}[h]
\centering
\caption{Distillation results with different ratios of the combination with hard labels and soft labels. The teacher network is ResNet-50 and the student is ResNet-18.}
\label{tab:long-table_hard}
\vspace{-0.13in}
\resizebox{.55\textwidth}{!}{%
\begin{tabular}{ccc}
 \toprule[1.1pt]
                            &                    & ImageNet                \\ \hline
Ratio (hard label -- soft label)    & w/ LS              & Acc. (Top1/Top5)   \\
 &  {\ding{56}} &71.592/90.386   \\
                          
 \rowcolor{mygray} \cellcolor{white}   \multirow{-2}{*}{0.3 -- 0.5}    &  {\ding{52}} & \bf  71.752/90.412 \\
                             \hline
 & {\ding{56}} & 71.484/90.218    \\
                           
 \rowcolor{mygray} \cellcolor{white}   \multirow{-2}{*}{0.5 -- 0.5}     &  {\ding{52}} & \bf  71.748/90.454  \\
                             \hline
 & {\ding{56}} & 71.164/90.196    \\
                           
  \rowcolor{mygray} \cellcolor{white}  \multirow{-2}{*}{0.7 -- 0.3}    &  {\ding{52}} & \bf  71.314/90.200  \\
\bottomrule[1.1pt]                 
\end{tabular}
}
\end{table}

\section{Results of Knowledge Distillation with Hard and Soft Labels} \label{hard}
Here we verify whether label smoothing is still effective when adopting hard labels in knowledge distillation. As we mentioned above, the traditional distillation loss can be formulated as $\
\lambda\mathcal{H}(\bm p^{{\mathcal S}_\mathbf{w}}, \bm y)+(1-\lambda)\mathcal{H}(\bm p^{{\mathcal S}_\mathbf{w}}/\mathcal{T}, \bm p^{\mathcal{T}_\mathbf{w}}/\mathcal{T})$. We use ResNet-50 as the teacher and ResNet-18 as the student and conduct experiments with three different ratios of $\lambda$: 0.3, 0.5, 0.7. Our Top-1/5 results are given in Table~\ref{tab:long-table_hard}. We can see the teacher networks with label smoothing still distills better students than the teacher without label smoothing. Also, with a higher probability of hard labels, the performance declines.

\section{Minor Probabilities in Teacher Network} \label{minor}
The visualization of minor probabilities is shown in Fig.~\ref{fig:weight_supervision} (2), we have two interesting observations in it: 1) In the enlarged area of Fig.~\ref{fig:weight_supervision} (2), we can observe that several classes have values are close to zero, this phenomenon means in ImageNet dataset, there are several categories that are very unique to other classes and the model will barely predict other classes to them; 2) We can also observe more ``red'' areas gather at the bottom of the bars. It reflects that the probabilities from the model trained with label smoothing will be assigned to more classes with smaller values, this is also consistent with what the label smoothing operation actually does.

\section{Supplementary Metrics of Inter-class/Entire Variations} \label{inter}
To better understand the behavior of erased information by label smoothing across different classes, we also introduce the inter-class stability metric as follows:
\begin{equation}
\mathcal{S}^\textrm{inter}_{\textrm{Stability}}=1-\frac{1}{\bm K}\sum_{c=1}^{\bm K}(||  \overline {\bm p}_{\{c\}}^{\mathcal T_\mathbf{w}}- \overline {\bm p}_{\{\bm {all}\}}^{\mathcal T_\mathbf{w}}||^{2})
\end{equation}
where $\bm K$ is the number of classes. $\overline {\bm p}_{\{\bm {all}\}}^{\mathcal T_\mathbf{w}}$ is average of probability across the entire dataset. 
$\overline {\bm p}_{\{c\}}^{\mathcal T_\mathbf{w}}$ is the mean of $\bm p^{\mathcal T_\mathbf{w}}$ in class $c$. $\mathcal{S}^\textrm{inter}_{\textrm{Stability}}$ can be regarded as a supplement for our intra-class stability metric to measure the inter-class stability. The average probability within each class $\overline {\bm p}_{\{c\}}^{\mathcal T_\mathbf{w}}$ is defined as:
\begin{equation}
\overline {\bm p}_{\{c\}}^{\mathcal T_\mathbf{w}}=\frac{1}{\bm n_c}\sum_{i=1}^{\bm n_c}  {\bm p}_{\{i,c\}}^{\mathcal T_\mathbf{w}}
\end{equation}
where  $i$ is the index of images and $\bm n_c$ is the \#image in class $c$. The average probability across entire dataset $\overline {\bm p}_{\{\bm {all}\}}^{\mathcal T_\mathbf{w}}$ is defined as:
\begin{equation}
\overline {\bm p}_{\{\bm {all}\}}^{\mathcal T_\mathbf{w}}=\frac{1}{\bm N}\sum_{i=1}^{\bm N}  {\bm p}_{\{i\}}^{\mathcal T_\mathbf{w}}
\end{equation}
where $\bm N$ is the number of samples in the entire dataset. This metric utilizes the probabilities of inter-class variance to measure the stability of a teacher’s prediction.  
The results on various network architectures are shown in Table~\ref{tab:stability-table_inter}.

\renewcommand{\arraystretch}{1.2}
\setlength{\tabcolsep}{0.5em}
\begin{table}[t]
\centering
\caption{Accuracy and inter-class stability results with and without label smoothing on ImageNet-1K. Here we show (1-$\mathcal{S}^\textbf{inter}_\textbf{Stability}$), which denotes the aggregated inter-class variance (the lower the better). Red numbers are the quantitative values of the erased information by label smoothing.}
\label{tab:stability-table_inter}
\vspace{-0.1in}
\resizebox{.99\textwidth}{!}{%
\begin{tabular}{lcc|cc}
 \toprule[1.1pt]
      \bf Netowrks             &\bf Acc. (\%) w/o LS     & \bf (1-$\mathcal{S}^\textrm{inter}_\textbf{Stability}$) w/o LS  & \bf Acc. (\%) w/ LS    &\bf (1-$\mathcal{S}^\textrm{inter}_\textbf{Stability}$) w/ LS       \\ \hline
{ResNet-18}~\citep{he2016deep}  & 69.758/89.078 & 0.1858    & \bf 69.774/89.122 & \bf 0.1724 (\color{red} -0.0134) \\ 
{ResNet-50}~\citep{he2016deep}  &75.888/92.642 & 0.1733    & \bf 76.130/92.936 & \bf 0.1610 (\color{red}-0.0123) \\ 
{ResNet-101}~\citep{he2016deep} & 77.374/93.546 &0.1671    & \bf 77.726/93.830 & \bf 0.1646 (\color{red}-0.0025) \\ \hline
{MobileNet v2}~\citep{sandler2018mobilenetv2}  & 71.878/90.286& 0.1797  & --  &\bf 0.1726 (\color{red}-0.0071) \\ \hline
{DenseNet121}~\citep{huang2017densely}  &  74.434/91.972 & 0.1763    & --   & \bf 0.1666 (\color{red}-0.0097) \\ \hline
{ResNeXt50 32$\times$4d}~\citep{xie2017aggregated}  & 77.618/93.698&\bf 0.1658   & \bf 77.774/93.642 &  0.1729 (\color{blue}+0.0071) \\ \hline
{Wide ResNet50}~\citep{zagoruyko2016wide}  & \bf 78.468/94.086 & \bf 0.1602  &  77.808/93.682 &  0.1688 (\color{blue}+0.0086) \\ \hline
{ResNeXt101 32$\times$8d}~\citep{xie2017aggregated}  & 79.312/94.526 &\bf 0.1596    & \bf 79.698/94.768 &  0.1677 (\color{blue}+0.0081) \\ 
\bottomrule[1.1pt]                 
\end{tabular}
}
\vspace{-0.1in}
\end{table}

\end{document}

%% file: math_commands.tex
\usepackage{amsmath,amsfonts,bm}

\def\eqref#1{equation~\ref{#1}}
\def\1{\bm{1}}

\DeclareMathAlphabet{\mathsfit}{\encodingdefault}{\sfdefault}{m}{sl}
\SetMathAlphabet{\mathsfit}{bold}{\encodingdefault}{\sfdefault}{bx}{n}

%% file: camera-ready.bbl
\begin{thebibliography}{42}
\providecommand{\natexlab}[1]{#1}
\providecommand{\url}[1]{\texttt{#1}}
\expandafter\ifx\csname urlstyle\endcsname\relax
  \providecommand{\doi}[1]{doi: #1}\else
  \providecommand{\doi}{doi: \begingroup \urlstyle{rm}\Url}\fi

\bibitem[Arani et~al.(2019)Arani, Sarfraz, and Zonooz]{arani2019improving}
Elahe Arani, Fahad Sarfraz, and Bahram Zonooz.
\newblock Improving generalization and robustness with noisy collaboration in
  knowledge distillation.
\newblock \emph{arXiv preprint arXiv:1910.05057}, 2019.

\bibitem[Bagherinezhad et~al.(2018)Bagherinezhad, Horton, Rastegari, and
  Farhadi]{bagherinezhad2018label}
Hessam Bagherinezhad, Maxwell Horton, Mohammad Rastegari, and Ali Farhadi.
\newblock Label refinery: Improving imagenet classification through label
  progression.
\newblock \emph{arXiv preprint arXiv:1805.02641}, 2018.

\bibitem[Chiu et~al.(2018)Chiu, Sainath, Wu, Prabhavalkar, Nguyen, Chen,
  Kannan, Weiss, Rao, Gonina, et~al.]{chiu2018state}
Chung-Cheng Chiu, Tara~N Sainath, Yonghui Wu, Rohit Prabhavalkar, Patrick
  Nguyen, Zhifeng Chen, Anjuli Kannan, Ron~J Weiss, Kanishka Rao, Ekaterina
  Gonina, et~al.
\newblock State-of-the-art speech recognition with sequence-to-sequence models.
\newblock In \emph{2018 IEEE International Conference on Acoustics, Speech and
  Signal Processing (ICASSP)}, pp.\  4774--4778. IEEE, 2018.

\bibitem[Chorowski \& Jaitly(2017)Chorowski and Jaitly]{chorowski2017towards}
Jan Chorowski and Navdeep Jaitly.
\newblock Towards better decoding and language model integration in sequence to
  sequence models.
\newblock \emph{Proc. Interspeech 2017}, pp.\  523--527, 2017.

\bibitem[Deng et~al.(2009)Deng, Dong, Socher, Li, Li, and
  Fei-Fei]{deng2009imagenet}
Jia Deng, Wei Dong, Richard Socher, Li-Jia Li, Kai Li, and Li~Fei-Fei.
\newblock Imagenet: A large-scale hierarchical image database.
\newblock In \emph{2009 IEEE conference on computer vision and pattern
  recognition}, pp.\  248--255, 2009.

\bibitem[Goyal et~al.(2017)Goyal, Doll{\'a}r, Girshick, Noordhuis, Wesolowski,
  Kyrola, Tulloch, Jia, and He]{goyal2017accurate}
Priya Goyal, Piotr Doll{\'a}r, Ross Girshick, Pieter Noordhuis, Lukasz
  Wesolowski, Aapo Kyrola, Andrew Tulloch, Yangqing Jia, and Kaiming He.
\newblock Accurate, large minibatch sgd: Training imagenet in 1 hour.
\newblock \emph{arXiv preprint arXiv:1706.02677}, 2017.

\bibitem[He et~al.(2016)He, Zhang, Ren, and Sun]{he2016deep}
Kaiming He, Xiangyu Zhang, Shaoqing Ren, and Jian Sun.
\newblock Deep residual learning for image recognition.
\newblock In \emph{Proceedings of the IEEE conference on computer vision and
  pattern recognition}, pp.\  770--778, 2016.

\bibitem[He et~al.(2019)He, Zhang, Zhang, Zhang, Xie, and Li]{he2019bag}
Tong He, Zhi Zhang, Hang Zhang, Zhongyue Zhang, Junyuan Xie, and Mu~Li.
\newblock Bag of tricks for image classification with convolutional neural
  networks.
\newblock In \emph{Proceedings of the IEEE Conference on Computer Vision and
  Pattern Recognition}, pp.\  558--567, 2019.

\bibitem[Hinton et~al.(2015)Hinton, Vinyals, and Dean]{hinton2015distilling}
Geoffrey Hinton, Oriol Vinyals, and Jeff Dean.
\newblock Distilling the knowledge in a neural network.
\newblock \emph{arXiv preprint arXiv:1503.02531}, 2015.

\bibitem[Huang et~al.(2017)Huang, Liu, Van Der~Maaten, and
  Weinberger]{huang2017densely}
Gao Huang, Zhuang Liu, Laurens Van Der~Maaten, and Kilian~Q Weinberger.
\newblock Densely connected convolutional networks.
\newblock In \emph{Proceedings of the IEEE conference on computer vision and
  pattern recognition}, pp.\  4700--4708, 2017.

\bibitem[Khosla et~al.(2020)Khosla, Teterwak, Wang, Sarna, Tian, Isola,
  Maschinot, Liu, and Krishnan]{khosla2020supervised}
Prannay Khosla, Piotr Teterwak, Chen Wang, Aaron Sarna, Yonglong Tian, Phillip
  Isola, Aaron Maschinot, Ce~Liu, and Dilip Krishnan.
\newblock Supervised contrastive learning.
\newblock \emph{arXiv preprint arXiv:2004.11362}, 2020.

\bibitem[Kingma \& Ba(2014)Kingma and Ba]{kingma2014adam}
Diederik~P Kingma and Jimmy Ba.
\newblock Adam: A method for stochastic optimization.
\newblock \emph{arXiv preprint arXiv:1412.6980}, 2014.

\bibitem[Liu et~al.(2018)Liu, Wu, Luo, Yang, Liu, and Cheng]{liu2018bi}
Zechun Liu, Baoyuan Wu, Wenhan Luo, Xin Yang, Wei Liu, and Kwang-Ting Cheng.
\newblock Bi-real net: Enhancing the performance of 1-bit cnns with improved
  representational capability and advanced training algorithm.
\newblock In \emph{Proceedings of the European conference on computer vision
  (ECCV)}, pp.\  722--737, 2018.

\bibitem[Liu et~al.(2020)Liu, Shen, Savvides, and Cheng]{liu2020reactnet}
Zechun Liu, Zhiqiang Shen, Marios Savvides, and Kwang-Ting Cheng.
\newblock Reactnet: Towards precise binary neural network with generalized
  activation functions.
\newblock \emph{arXiv preprint arXiv:2003.03488}, 2020.

\bibitem[Liu et~al.(2019)Liu, Miao, Zhan, Wang, Gong, and Yu]{liu2019large}
Ziwei Liu, Zhongqi Miao, Xiaohang Zhan, Jiayun Wang, Boqing Gong, and Stella~X
  Yu.
\newblock Large-scale long-tailed recognition in an open world.
\newblock In \emph{Proceedings of the IEEE Conference on Computer Vision and
  Pattern Recognition}, pp.\  2537--2546, 2019.

\bibitem[Lukasik et~al.(2020)Lukasik, Bhojanapalli, Menon, and
  Kumar]{lukasik2020does}
Michal Lukasik, Srinadh Bhojanapalli, Aditya~Krishna Menon, and Sanjiv Kumar.
\newblock Does label smoothing mitigate label noise?
\newblock \emph{arXiv preprint arXiv:2003.02819}, 2020.

\bibitem[Martinez et~al.(2020)Martinez, Yang, Bulat, and
  Tzimiropoulos]{Martinez2020Training}
Brais Martinez, Jing Yang, Adrian Bulat, and Georgios Tzimiropoulos.
\newblock Training binary neural networks with real-to-binary convolutions.
\newblock In \emph{International Conference on Learning Representations}, 2020.

\bibitem[Mghabbar \& Ratnamogan(2020)Mghabbar and
  Ratnamogan]{mghabbar2020building}
Idriss Mghabbar and Pirashanth Ratnamogan.
\newblock Building a multi-domain neural machine translation model using
  knowledge distillation.
\newblock \emph{arXiv preprint arXiv:2004.07324}, 2020.

\bibitem[M{\"u}ller et~al.(2019)M{\"u}ller, Kornblith, and
  Hinton]{muller2019does}
Rafael M{\"u}ller, Simon Kornblith, and Geoffrey~E Hinton.
\newblock When does label smoothing help?
\newblock In \emph{Advances in Neural Information Processing Systems}, pp.\
  4696--4705, 2019.

\bibitem[Pereyra et~al.(2017)Pereyra, Tucker, Chorowski, Kaiser, and
  Hinton]{pereyra2017regularizing}
Gabriel Pereyra, George Tucker, Jan Chorowski, {\L}ukasz Kaiser, and Geoffrey
  Hinton.
\newblock Regularizing neural networks by penalizing confident output
  distributions.
\newblock \emph{arXiv preprint arXiv:1701.06548}, 2017.

\bibitem[Rastegari et~al.(2016)Rastegari, Ordonez, Redmon, and
  Farhadi]{rastegari2016xnor}
Mohammad Rastegari, Vicente Ordonez, Joseph Redmon, and Ali Farhadi.
\newblock Xnor-net: Imagenet classification using binary convolutional neural
  networks.
\newblock In \emph{European conference on computer vision}, pp.\  525--542,
  2016.

\bibitem[Reed(2001)]{William2001}
William~J. Reed.
\newblock The pareto, zipf and other power laws.
\newblock \emph{Economics Letters}, 2001.

\bibitem[Romero et~al.(2015)Romero, Ballas, Kahou, Chassang, Gatta, and
  Bengio]{romero2014fitnets}
Adriana Romero, Nicolas Ballas, Samira~Ebrahimi Kahou, Antoine Chassang, Carlo
  Gatta, and Yoshua Bengio.
\newblock Fitnets: Hints for thin deep nets.
\newblock In \emph{ICLR}, 2015.

\bibitem[Sandler et~al.(2018)Sandler, Howard, Zhu, Zhmoginov, and
  Chen]{sandler2018mobilenetv2}
Mark Sandler, Andrew Howard, Menglong Zhu, Andrey Zhmoginov, and Liang-Chieh
  Chen.
\newblock Mobilenetv2: Inverted residuals and linear bottlenecks.
\newblock In \emph{Proceedings of the IEEE conference on computer vision and
  pattern recognition}, pp.\  4510--4520, 2018.

\bibitem[Shen \& Savvides(2020)Shen and Savvides]{shen2020meal}
Zhiqiang Shen and Marios Savvides.
\newblock Meal v2: Boosting vanilla resnet-50 to 80\%+ top-1 accuracy on
  imagenet without tricks.
\newblock \emph{arXiv preprint arXiv:2009.08453}, 2020.

\bibitem[Shen et~al.(2019)Shen, He, and Xue]{shen2019meal}
Zhiqiang Shen, Zhankui He, and Xiangyang Xue.
\newblock Meal: Multi-model ensemble via adversarial learning.
\newblock In \emph{Proceedings of the AAAI Conference on Artificial
  Intelligence}, volume~33, pp.\  4886--4893, 2019.

\bibitem[Shen et~al.(2020)Shen, Liu, Liu, Savvides, Darrell, and
  Xing]{shen2020mix}
Zhiqiang Shen, Zechun Liu, Zhuang Liu, Marios Savvides, Trevor Darrell, and
  Eric Xing.
\newblock Un-mix: Rethinking image mixtures for unsupervised visual
  representation learning.
\newblock \emph{arXiv preprint arXiv:2003.05438}, 2020.

\bibitem[Szegedy et~al.(2016)Szegedy, Vanhoucke, Ioffe, Shlens, and
  Wojna]{szegedy2016rethinking}
Christian Szegedy, Vincent Vanhoucke, Sergey Ioffe, Jon Shlens, and Zbigniew
  Wojna.
\newblock Rethinking the inception architecture for computer vision.
\newblock In \emph{Proceedings of the IEEE conference on computer vision and
  pattern recognition}, pp.\  2818--2826, 2016.

\bibitem[Tan et~al.(2019)Tan, Ren, He, Qin, and Liu]{tan2018multilingual}
Xu~Tan, Yi~Ren, Di~He, Tao Qin, and Tie-Yan Liu.
\newblock Multilingual neural machine translation with knowledge distillation.
\newblock In \emph{International Conference on Learning Representations}, 2019.

\bibitem[Tang et~al.(2020)Tang, Shivanna, Zhao, Lin, Singh, Chi, and
  Jain]{tang2020understanding}
Jiaxi Tang, Rakesh Shivanna, Zhe Zhao, Dong Lin, Anima Singh, Ed~H Chi, and
  Sagar Jain.
\newblock Understanding and improving knowledge distillation.
\newblock \emph{arXiv preprint arXiv:2002.03532}, 2020.

\bibitem[Touvron et~al.(2019)Touvron, Vedaldi, Douze, and
  J{\'e}gou]{touvron2019FixRes}
Hugo Touvron, Andrea Vedaldi, Matthijs Douze, and Herv{\'e} J{\'e}gou.
\newblock Fixing the train-test resolution discrepancy.
\newblock In \emph{Advances in Neural Information Processing Systems
  (NeurIPS)}, 2019.

\bibitem[Van~Horn et~al.(2018)Van~Horn, Mac~Aodha, Song, Cui, Sun, Shepard,
  Adam, Perona, and Belongie]{van2018inaturalist}
Grant Van~Horn, Oisin Mac~Aodha, Yang Song, Yin Cui, Chen Sun, Alex Shepard,
  Hartwig Adam, Pietro Perona, and Serge Belongie.
\newblock The inaturalist species classification and detection dataset.
\newblock In \emph{Proceedings of the IEEE conference on computer vision and
  pattern recognition}, pp.\  8769--8778, 2018.

\bibitem[Vaswani et~al.(2017)Vaswani, Shazeer, Parmar, Uszkoreit, Jones, Gomez,
  Kaiser, and Polosukhin]{vaswani2017attention}
Ashish Vaswani, Noam Shazeer, Niki Parmar, Jakob Uszkoreit, Llion Jones,
  Aidan~N Gomez, {\L}ukasz Kaiser, and Illia Polosukhin.
\newblock Attention is all you need.
\newblock In \emph{Advances in neural information processing systems}, pp.\
  5998--6008, 2017.

\bibitem[Wah et~al.(2011{\natexlab{a}})Wah, Branson, Welinder, Perona, and
  Belongie]{WahCUB_200_2011}
C.~Wah, S.~Branson, P.~Welinder, P.~Perona, and S.~Belongie.
\newblock {The Caltech-UCSD Birds-200-2011 Dataset}.
\newblock Technical report, 2011{\natexlab{a}}.

\bibitem[Wah et~al.(2011{\natexlab{b}})Wah, Branson, Welinder, Perona, and
  Belongie]{wah2011caltech}
Catherine Wah, Steve Branson, Peter Welinder, Pietro Perona, and Serge
  Belongie.
\newblock The caltech-ucsd birds-200-2011 dataset.
\newblock 2011{\natexlab{b}}.

\bibitem[Xie et~al.(2019)Xie, Hovy, Luong, and Le]{xie2019self}
Qizhe Xie, Eduard Hovy, Minh-Thang Luong, and Quoc~V Le.
\newblock Self-training with noisy student improves imagenet classification.
\newblock \emph{arXiv preprint arXiv:1911.04252}, 2019.

\bibitem[Xie et~al.(2017)Xie, Girshick, Doll{\'a}r, Tu, and
  He]{xie2017aggregated}
Saining Xie, Ross Girshick, Piotr Doll{\'a}r, Zhuowen Tu, and Kaiming He.
\newblock Aggregated residual transformations for deep neural networks.
\newblock In \emph{Proceedings of the IEEE conference on computer vision and
  pattern recognition}, pp.\  1492--1500, 2017.

\bibitem[Yuan et~al.(2019)Yuan, Tay, Li, Wang, and Feng]{yuan2019revisit}
Li~Yuan, Francis~EH Tay, Guilin Li, Tao Wang, and Jiashi Feng.
\newblock Revisit knowledge distillation: a teacher-free framework.
\newblock \emph{arXiv preprint arXiv:1909.11723}, 2019.

\bibitem[Yun et~al.(2019)Yun, Han, Oh, Chun, Choe, and Yoo]{yun2019cutmix}
Sangdoo Yun, Dongyoon Han, Seong~Joon Oh, Sanghyuk Chun, Junsuk Choe, and
  Youngjoon Yoo.
\newblock Cutmix: Regularization strategy to train strong classifiers with
  localizable features.
\newblock In \emph{Proceedings of the IEEE International Conference on Computer
  Vision}, pp.\  6023--6032, 2019.

\bibitem[Zagoruyko \& Komodakis(2016)Zagoruyko and
  Komodakis]{zagoruyko2016wide}
Sergey Zagoruyko and Nikos Komodakis.
\newblock Wide residual networks.
\newblock \emph{arXiv preprint arXiv:1605.07146}, 2016.

\bibitem[Zhou et~al.(2017)Zhou, Lapedriza, Khosla, Oliva, and
  Torralba]{zhou2017places}
Bolei Zhou, Agata Lapedriza, Aditya Khosla, Aude Oliva, and Antonio Torralba.
\newblock Places: A 10 million image database for scene recognition.
\newblock \emph{IEEE Transactions on Pattern Analysis and Machine
  Intelligence}, 2017.

\bibitem[Zhou et~al.(2020)Zhou, Gu, and Neubig]{Zhou2020Understanding}
Chunting Zhou, Jiatao Gu, and Graham Neubig.
\newblock Understanding knowledge distillation in non-autoregressive machine
  translation.
\newblock In \emph{International Conference on Learning Representations}, 2020.

\end{thebibliography}
